\documentclass[final]{cvpr}

\usepackage{times}
\usepackage{epsfig}
\usepackage{amsmath}
\usepackage{amssymb}
\usepackage[utf8]{inputenc} 
\usepackage[T1]{fontenc}    
\usepackage{url}            
\usepackage{booktabs}       
\usepackage{amsfonts}       
\usepackage{nicefrac}       
\usepackage{microtype}      
\usepackage{graphicx,subcaption}
\usepackage[bottom]{footmisc}
\usepackage[ruled]{algorithm2e}

\usepackage[pagebackref=true,breaklinks=true,colorlinks,bookmarks=false]{hyperref}

\def\cvprPaperID{2835} 
\def\confYear{CVPR 2021}

\begin{document}

\title{Task-Aware Variational Adversarial Active Learning}

\author{Kwanyoung Kim$^{1,4}$, Dongwon Park$^1$, Kwang In Kim$^{2,3}$, Se Young Chun$^{1,3,*}$\\
$^1$Department of Electrical Engineering, $^2$Department of Computer Science and Engineering,\\
$^3$Graduate School of Artificial Intelligence, UNIST, Republic of Korea.
{\tt\small $^*$sychun@unist.ac.kr}
}

\maketitle

\newcommand{\tnote}[3]{{\color{#2}#1: #3}}
\newcommand{\kimki}[1]{\tnote{[KIMKI}{magenta}{#1}]}
\newcommand{\sychun}[1]{\tnote{[SYCHUN}{green}{#1}]}
\newcommand{\kwanyoung}[1]{\tnote{[KWAN}{red}{#1}]}

\begin{abstract}
Often, labeling large amount of data is challenging due to high labeling cost limiting the application domain of deep learning techniques. Active learning (AL) tackles this by querying the most informative samples to be annotated among unlabeled pool. Two promising directions for AL that have been recently explored are task-agnostic approach to select data points that are far from the current labeled pool and task-aware approach that relies on the perspective of task model. Unfortunately, the former does not exploit structures from tasks and the latter does not seem to well-utilize overall data distribution. Here, we propose task-aware variational adversarial AL (TA-VAAL) that modifies task-agnostic VAAL, that considered data distribution of both label and unlabeled pools, by relaxing task learning loss prediction to ranking loss prediction and by using ranking conditional generative adversarial network to embed normalized ranking loss information on VAAL. Our proposed TA-VAAL outperforms state-of-the-arts on various benchmark datasets for classifications with balanced / imbalanced labels as well as semantic segmentation and its task-aware and task-agnostic AL properties were confirmed with our in-depth analyses.
\end{abstract}

\begin{figure*}[!t]
	\centering	
	\includegraphics[width=1\linewidth]{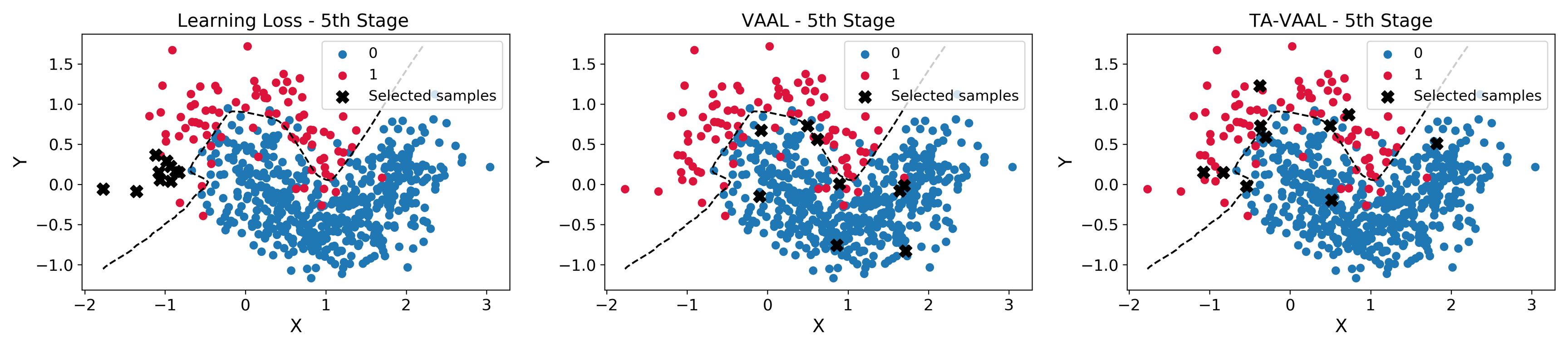}  
	\vspace{-2em}
	\caption{Visual results of active learning methods (Learning loss~\cite{yoo2019learning}, VAAL~\cite{sinha2019variational}, our TA-VAAL) on imbalanced toy example at the 5th stage. \emph{Red} and \emph{blue} dots indicate samples assigned to class 0 and 1, respectively. Ten samples at that stage (denoted by \emph{black} cross) were selected using each method. The oracle decision boundary of the model is shown as a black dash line.
	Learning loss identified difficult samples near the decision boundary and VAAL found influential samples over the entire set. Our TA-VAAL selected samples that are both difficult (near decision boundary) and influential (over the entire set).} 
	\label{fig:Toy}
		\vspace{-1em}
\end{figure*}

\section{Introduction}

\footnotetext[4]{Currently with KAIST. This work was done when he was with UNIST.}

Deep learning has achieved remarkable performance in various computer vision tasks 
such as classification~\cite{krizhevsky2012imagenet,he2016deep}, 
object detection~\cite{ren2015faster,redmon2016you}, and semantic segmentation~\cite{long2015fully,chen2018encoder} 
due to massive datasets with annotations such as ImageNet for image classification~\cite{imagenet_cvpr09} 
and PASCAL VOC for classification, detection, segmentation~\cite{Everingham10}.
Obtaining good annotations is challenging and has often been a large-scale project.
Moreover, there are often cases where labeling massive amount of data is even more challenging or  infeasible 
due to high labeling cost such as labeling by experts~\cite{esteva2017dermatologist}
or long labeling time per large-scale sample such as videos~\cite{Haija2016youtube8m} or 
pathology images~\cite{campanella2019clinical}.
Labeling cost seems to be a factor to limit the scope of applicability of deep learning 
to more research areas and more institutes with less labeling budget.

Active learning (AL) is one of the approaches to 
overcoming limited labeling budget 
by selecting data to label for the best possible performance~\cite{settles2009active,gal2017deep}.
AL has been widely investigated 
in relatively traditional machine learning 
settings~\cite{cohn1996active,tong2001support,brinker2003incorporating,melville2004diverse,nguyen2004active,settles2009active,houlsby2011bayesian,mac2014hierarchical,wang2015querying,sundin2019active,pinsler2019bayesian}
and recently 
in deep learning settings~\cite{gal2017deep,sener2017active,yang2017suggestive,yoo2019learning,tran2019bayesian,sinha2019variational,kirsch2019batchbald}.

Existing AL approaches can be categorized into two groups: Task-agnostic (or distribution-based) and task-aware methods. Suppose that our goal is to learn a functional model $f$ that maps from the input domain $\mathcal{X}$ to the corresponding output domain $\mathcal{Y}$, each equipped with the corresponding probability distributions $P(x)$ and $P(y)$. Task-agnostic approaches select data instances to label by exploiting the input distribution $P(x)$. These are especially effective in identifying \emph{influential} points, \eg these lying in high-density regions such that once labeled, large numbers of \emph{neighboring} samples can benefit from propagating these labels~\cite{mac2014hierarchical,yang2015multi,sener2017active,sinha2019variational}. A major drawback of these approaches is that they do not take account how outputs $y$ depend on inputs $x$: For example, for classifications, it would be more effective to label data instances that lie in the vicinity of decision boundaries than these lying in high-density regions where most data points belong to the same class.

Task-aware approaches explicitly address this limitation by modeling such dependence, \eg via estimating the conditional distribution $P(y|x)$. These are effective in identifying \emph{difficult} data points (\eg these close to decision boundaries)~\cite{wang2015querying,houlsby2011bayesian,gal2017deep,tran2019bayesian,yoo2019learning}. However, they do not directly consider how the labeled samples make influence on the entire dataset. Further, as $P(y|x)$ is unknown a priori, the label selection process has to rely on the learner $f$ as a surrogate to $P(y|x)$ but such a learner might be inaccurate at the early stage of AL, thereby providing a poor estimate of $P(y|x)$. 

Recently, there was an attempt (SRAAL) to combine the task-aware and task-agnostic approaches with a uncertainty indicator and with a unified representation for both labeled and unlabeled data~\cite{zhang2020state}. Even though SRAAL achieved state-of-the-art performance, it did not use the information about the loss that is directly related to the given task~\cite{yoo2019learning}
and its task learner seems to be limited only to VAE-type networks with a latent space for its unified representation. Moreover, its implementation is not yet available online.

In this paper, we propose a novel alternative AL scheme that combines the benefits of these two groups of approaches. Specifically, our algorithm builds upon two recent state-of-the-art approaches: \emph{Variational adversarial active learning} (VAAL)~\cite{sinha2019variational} models how adding labels to selected data points make influence on the entire set. As a model-agnostic approach, this method does not exploit the structure $P(y|x)$ of the problem at hand. We address this by combining it with the recent \emph{learning loss} approach~\cite{yoo2019learning}. This algorithm learns to estimate the errors of the predictions (loss) made by the learner and therefore helps identify difficult data points.

Here is the summary of our contributions:\\
$\bullet$ Proposing to relax the goal of loss prediction module~\cite{yoo2019learning} from accurate loss prediction to loss \emph{ranking} prediction, which is still directly connected to the task. This relaxation leads to altering the loss for learning prediction module to remove margins for ranking and to add ranking loss in~\cite{saquil2018ranking}.\\ 
$\bullet$ Proposing \emph{Task-Aware Variational Adversarial Active Learning} (TA-VAAL) to embed the normalized ranking loss information from any given task learner (with or without latent space) on the latent space of VAAL~\cite{sinha2019variational} via ranking conditional generative adversarial network (RankCGAN)~\cite{saquil2018ranking} to reshape the latent space of it.
This approach is significantly more robust than the original learning loss approach, especially at the early stage. 
By combining these two algorithms with our embedding strategy, our method offers the capability of identifying \emph{difficult} and \emph{influential} data points (see Figure~\ref{fig:Toy}; see Section~\ref{s:experiments} for details).\\
$\bullet$ Demonstrating the superior performance of our proposed TA-VAAL over state-of-the-art works (Learning loss~\cite{yoo2019learning}, VAAL~\cite{sinha2019variational}, Coreset~\cite{sener2017active}, Monte-Carlo dropout~\cite{gal2017deep}) by evaluating on various classification benchmark datasets: CIFAR10, CIFAR100 that have the same number of images per class (balanced) as well as Caltech101, modified CIFAR10 that has different numbers of images for classes (imbalanced), and on Cityscapes semantic segmentation benchmark dataset and by in-depth empirical analyses to confirm our proposed approach.
Our codes will be available online. 

\section{Related Works}
\label{s:relatedwork}
There have been a number of AL works to select the most informative samples and we categorized them into two approaches: task-aware (or model uncertainty-based) and task-agnostic approaches.
The former is using unlabeled data in a passive way while the latter is using unlabeled data in an active way. 
In other words, the former has sample selection rules that are not affected by unlabeled data, but simply are applied to it, while
the latter exploits both labeled and unlabeled data to build up sample selection rules (or train deep neural networks (DNNs) for them).

Task-aware approach defined and used metrics for sample selection with labeled data. 
For example, the minimum distance from decision boundaries (or classification hyperplanes)
can be used to select samples with the most ambiguous classification results~\cite{tong2001support,brinker2003incorporating}. 
Empirical risk is used to minimize an upper bound of true risk so that one can
query the most informative samples that are the most uncertain and representative~\cite{wang2015querying}.
Bayesian active learning by disagreement (BALD) maximizes the mutual information between model predictions and model parameters~\cite{houlsby2011bayesian}. 
Then, BALD was extended to accommodate DNNs with Bayesian neural network and Monte-Carlo dropout~\cite{gal2017deep}.
Bayesian generative active deep learning was proposed to utilize both labeled data and labeled fake data 
to train a classifier (or a task-learner) as well as a discriminator for real / fake images~\cite{tran2019bayesian}.
Even though~\cite{tran2019bayesian} uses deep generative models or VAEs, 
it does not use unlabeled data for training unlike our proposed TA-VAAL.
Yoo and Kweon~\cite{yoo2019learning} proposed an AL loss method that attaches ``loss prediction module'' to a task-learner.
The loss prediction module was trained to estimate target losses of unlabeled samples that were used as 
surrogates for model uncertainty based on 
feature information in mid-layers. 

Task-agnostic approach exploits both labeled and unlabeled data to form sample selection rules so that
selected samples are far from the distribution of labeled data and have the most well-representative information of unlabeled pool.
Clustering unlabeled data could help to choose samples from diverse clusters, not from one or small
number of clusters~\cite{nguyen2004active}. 
Expected error reduction using hierarchical clustering was developed for active sampling in a semi-supervised framework~\cite{mac2014hierarchical}.
An objective function with diversity constrain was proposed to impose diversity on the subset of data pool for multi-class AL~\cite{yang2015multi}.
Recently, there have also been works on 
task-agnostic AL with DNNs.
Core-set approach was proposed 
that minimizes the distance between labeled data and unlabeled data pool with intermediate feature information of trained convolutional DNN models~\cite{sener2017active}.
Gudovskiy~\textit{et al.}~\cite{gudovskiy2020deep} proposed to minimize distribution shift between unlabeled training set and weakly-labeled validation set for semi-supervised AL.
Sinha \textit{et al.}~\cite{sinha2019variational} proposed VAAL to train VAE that captures the representing information of both labeled and unlabeled data with adversarial learning to discriminate unlabeled samples from labeled data using the latent space information in the VAE. 

An extended version of VAAL (SRAAL)~\cite{zhang2020state} was proposed to combine task-aware and task-agnostic approaches with a uncertainty indicator and with a unified representation for both labeled and unlabeled data. However, SRAAL did not use the final information on task ($e.g.$, loss~\cite{yoo2019learning}), but used intermediate task information such as the latent space information from the task learner. Moreover, its task learner seems to be limited only to VAE-type networks with a latent space for its unified representation. 
In the meanwhile, our proposed TA-VAAL is a novel alternative to combine both task-aware and task-agnostic
approaches that
is another extension of task-agnostic VAAL to incorporate direct task related information (ranking loss) into the VAE framework. In addition, our TA-VAAL does not have any structural restriction for the task learner. We demonstrated that our proposed framework can accommodate both local task-related information and global data distribution structure so that high performance and reliability can be jointly achieved.

\section{Method}

Let us denote the pool of labeled data and annotations by $(X_L,Y_L)$ and the pool of unlabeled data by $X_U$.
The goal of AL is to select samples 
from $X_U$ with limited label budget,
to annotate them to yield pairs of sample / annotation $(x^{\ast},y^{\ast})$, and to add them to $(X_L,Y_L)$
for the best possible performance of a given task learner $T$ (DNN parametrized by $\theta_T$).
$(X_L,Y_L)$ will grow in size every stages. 
The task learner $T$ will be trained by minimizing the loss $\sum_{(x_L,y_L) \in (X_L,Y_L)} l_L$ at each stage where
$l_L = L_{T}(\hat{y}_L, y_L)$ is a loss value at $(x_L, y_L)$ and $\hat{y}_L = T(x_L)$ is a predicted label.

\subsection{Task loss prediction module as ``Ranker''}

\begin{figure}[!t]
	\centering
	\fbox{ \includegraphics[width=1\linewidth]{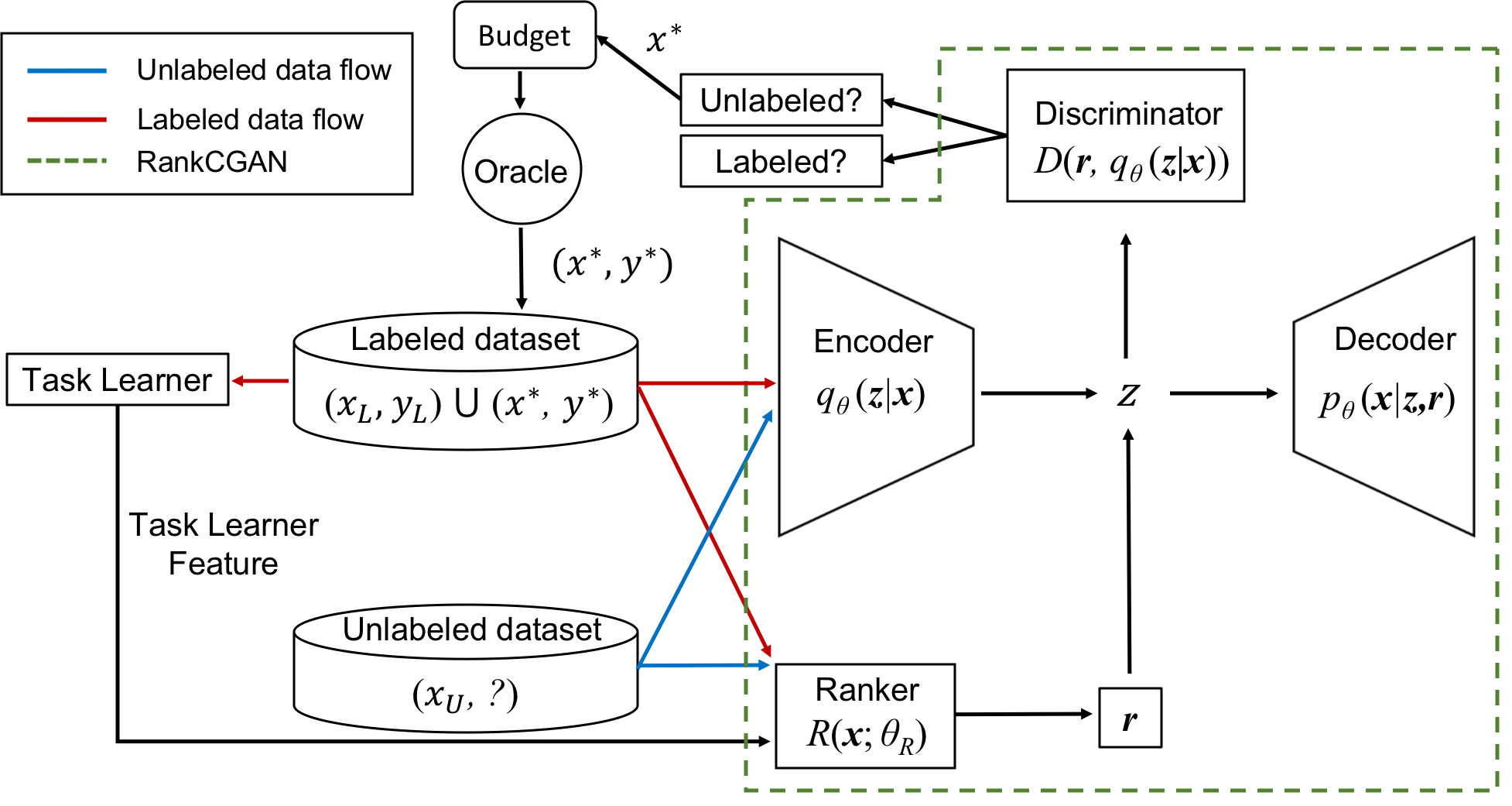} } 
	\vspace{-1em}
	\caption{A schematic diagram of our TA-VAAL: VAAL is effective at capturing the overall influence of labels propagated to the \emph{entire} distribution, but is agnostic to the nature of task at hand, \textit{i.e.}, VAAL is independent of the task labels predicted or provided as ground-truths. 
		By injecting the capability of capturing fine-grained 
		task label rank information, TA-VAAL helps focus on both influential and informative (or difficult) labels and adjust how they are propagated.}
	\label{fig:network}
	\vspace{-1em}
\end{figure}

Yoo and Kweon~\cite{yoo2019learning} proposed loss prediction module (LPM), denoted by $\Theta_{loss}$, to predict the loss value $\hat{l}_U = \Theta_{loss}(x_U)$ for $x_U \in X_U$ 
without ground truth labels.
LPM 
consists of global average pooling, fully connected layer and ReLU to predict 
unknown $l_U = L_T (T(x_U), y_U)$ 
where $y_U$ is unknown ground truth label (indicated as ``?'' in Figure~\ref{fig:network}). 
Since the task loss is usually decreasing over epochs, using the mean squared error (MSE) caused scaling issue. To avoid that, \cite{yoo2019learning} considered the difference between two losses 
and thus $(X_L, Y_L)$ was re-grouped into a set of pairs $(x_P, l_P) = \{ (x_i, x_j), (l_i, l_j) \}$.
Then, the loss for LPM is
\(
-({2}/{B}) \sum_{i=1}^{B/2} \max (0, -I_{i} \cdot (\hat{l}_i - \hat{l}_j) + \epsilon )) 
\)
where
$I_{i} = +1$ if $l_{i} > l_{j}$ and $-1$ otherwise,
and $\epsilon$ is a positive scalar that was set to $1$.
If there were $B$ elements in the original $(X_L, Y_L)$, this re-grouped set had $B/2$ elements of $(x_P, l_P)$.
This enabled LPM to be trained to yield accurate loss values. 

In this work, we relax the goal of LPM from predicting accurate loss values to estimating accurate ranking loss information. In other words, our proposed LPM will less care loss value itself, but more care relative loss rankings. For this purpose, we propose to exploit RankCGAN~\cite{saquil2018ranking} to connect between task-agnostic VAAL~\cite{sinha2019variational} and task-aware learning loss~\cite{yoo2019learning}.
While the LPM in~\cite{yoo2019learning} and
the concept of ``Ranker'' in RankCGAN~\cite{saquil2018ranking}
both utilized the difference between two predicted loss values in training losses, the former aimed to predict accurate losses and the latter focused on predicting ``ranking'' of the loss values.
Thus, our ranking loss is:
\begin{align}
	{L}_{R}(\hat{l}_P, l_{P})=-({2}/{B}) {\textstyle \sum_{i=1}^{B/2}} \{ I_{i}\log[\sigma(\hat{l}_i - \hat{l}_j)] \nonumber \\
	+ (1 - I_{i})\log[1-\sigma(\hat{l}_i - \hat{l}_j)] \}
	\label{eq:ranking_loss}
\end{align}
where $I_i = +1$ if $l_{i} > l_{j}$ and $0$ otherwise, $\hat{l}_i = R(x_i)$ with $R$ being the Ranker (DNN parametrized by $\theta_R$) that predicts loss, and $\sigma$ is the Sigmoid function. Rather than directly predicting \emph{loss} itself~\cite{yoo2019learning} (thus, the loss includes a margin $\epsilon$ to trust more on loss value itself and to emphasize less on preserving rankings), Ranker in our TA-VAAL is predicting relative \emph{rankings} of losses that can be embedded into the latent space of VAAL with the conditional latent variable $r$ with normalization via the Sigmoid function (thus, the rankings are strictly preserved with (\ref{eq:ranking_loss})). This choice has been motivated by 1) the observation that the relative comparisons of the target attributes are often easier to learn and predict than the absolute attribute values~\cite{saquil2018ranking}, 
and 2) for AL, ranking the data points to label is often sufficient~\cite{yoo2019learning}.
Moreover, while the learning loss in~\cite{yoo2019learning} is non-differentiable piecewise linear, our ranking loss (\ref{eq:ranking_loss}) is a smooth differentiable function that potentially has nice convergence properties for gradient based optimizations (see supplemental).

Finally, the total loss function of task learner with our proposed Ranker $R$ is expressed as
\begin{equation}
 L_{total} = L_{T}(\hat{y}_L,y_L) + \eta L_{R}(R(x_{P}),l_{P}) 
\label{eq:ranker_loss}
\end{equation}
where $\eta$ is a scaling parameter. 
We empirically found that 
training a task learner with this ranking loss (\ref{eq:ranking_loss}) 
was more stable and yielded better performance 
than the original learning loss~\cite{yoo2019learning} 
(see the ablation study in Section~\ref{sec:ablation}).

\subsection{Proposed task-aware VAAL (TA-VAAL)} 

Figure~\ref{fig:network} illustrates our proposed TA-VAAL that introduces Ranker, modifies the latent space of the original VAAL by incorporating a rank variable $r$ from Ranker, and inputs the normalized loss ranking information $r$ to both the decoder of VAAL and the discriminator of VAAL to select samples from unlabeled data pool.
Our proposed framework 
allows us to control the latent subspace with loss ranking predictions so that the overall latent space can be reshaped.

TA-VAAL is obtained with 
the following optimization:
\begin{align}
	\min_{q_{\theta}}\max_{D} \mathbb{E}_{z_{L} \sim p_{x_L}}[\log(D(r_L,q_{\theta}(z_L|x_L)))] \nonumber \\
	+ \mathbb{E}_{z_U \sim p_{x_U}}[\log(1-D(r_U,q_{\theta}(z_U|x_U)))]
	\label{eq:tavaalgan}
\end{align}
where $q_{\theta}$ is an encoder of the VAE, and 
$z_L$ and $z_U$ belong to the latent spaces for labeled and unlabeled data, respectively,
and $r_L$ and $r_U$ are the normalized outputs of the Ranker from labeled and unlabeled data, respectively.
$z_{L} \sim p_{x_L}$ implies $z_{L} = q_{\theta}(z_L | x_L)$ with $x_L \sim p_{data}$ and
$z_{U} \sim p_{x_U}$ is similar to $z_{L} \sim p_{x_L}$. 
By removing the rank information $r_L$ and $r_U$ from (\ref{eq:tavaalgan}), TA-VAAL boils down to 
VAAL that offers the capability of modeling the \emph{global} data distribution, but does not exploit the information gained from the task. Using the loss of the learner as a surrogate to such task information can help~\cite{yoo2019learning}, but assessing the \emph{individual} losses does not explicitly model how data instances make influence to each other and thus, they can be prone to noise and outliers. 

Here, we conjecture that task-related information further improves the overall performance of AL.
Our TA-VAAL bridges between model uncertainty-based and data distribution-based approaches 
in a \emph{tight} way 
by using conditional GAN (RankCGAN) so that the information about data distribution accounts for model uncertainty information (predicted loss ranking).
Our TA-VAAL will have an advantage to use more data (unlabeled data) over typical task-aware 
approach trained without unlabeled 
data and can offer significant improvements 
(see Section~\ref{s:experiments}).

\begin{algorithm}[!t]
	\caption{Training pipeline of our TA-VAAL}
	\label{algo-1}
	\SetKwInOut{Input}{Input}
	\SetKwInOut{Output}{Output}
	\SetKwInput{kwInit}{Initialize}
	\SetKwInput{kwset}{Given}
	\kwset{learning rates  $\zeta_1,\zeta_2,\zeta_3$, \# of epochs $N$;}
	\Input{labeled data ($x_{L},y_{L}$), unlabeled data $x_{U}$;}
	\kwInit{network parameters $\theta_{T},\theta_{R},\theta,\theta_{D}$;}
	\For {$n=1$ to N}
	{
		\lIf{$n = 1$}{$r_L, r_U \sim \mathcal{U}(0,1)$}
		\lElse{$r_L \gets R(x_L;\theta_{R})$ , $r_U \gets R(x_U; \theta_{R})$}	
		$L_{total} \gets L_{T} + \eta L_{R}$\\
		\lIf{$n \le 0.8 N$}{$\theta_{T} \gets \theta_{T} - \zeta_1 \nabla_{\theta_T} L_{total}$, $\;$ $\theta_{R} \gets \theta_{R} - \zeta_1 \nabla_{\theta_R}  L_{total}$}
		\lElse{$\theta_{T} \gets \theta_{T} - \zeta_1 \nabla_{\theta_T} L_{total}$}
		$L_{VAE} \gets L_{VAE}^{trans}+\lambda L_{VAE}^{adv}$,~$\theta \gets \theta - \zeta_2 \nabla_{\theta} L_{VAE}$\\
		$L_{D}$ using (\ref{eq:disc_loss}), $\;$ $\theta_{D} \gets \theta_{D} - \zeta_3 \nabla_{\theta_D} L_{D}$
	}
\end{algorithm}
		
\subsection{Training details for TA-VAAL}

The objective function ${L}_{VAE}^{trans}$ of the conditional VAE with ranking for learning features of both labeled and unlabeled pools can be formulated as
\begin{align}
	&\mathbb{E}[\log p_{\theta}(x_L|z_L,r_L)] - \beta \mathrm{KL}(q_{\theta}(z_L|x_L)||p_z)  \nonumber \\
	+&\mathbb{E}[\log p_{\theta}(x_U|z_U,r_U)] - \beta \mathrm{KL}(q_{\theta}(z_U|x_U)||p_z) 
	\label{eq:trans_loss}
\end{align}
where $q_{\theta}$ and $p_{\theta}$ are the encoder and decoder of the VAE, 
$p_z$ is Gaussian distribution, 
$\beta$ is a hyper-parameter, and
$\mathrm{KL}(\cdot)$ is Kullback-Leibler distance.
Another function for training 
is the conditional adversarial loss to represent both $q_{\theta}(z_L|x_L)$ and $q_{\theta}(z_U|x_U)$ with
the same distribution from labeled and unlabeled pools. 
The objective function ${L}_{VAE}^{adv}$ is
\begin{equation}
	-\mathbb{E}[\log D(r_{L},q_{\theta}(z_L|x_L))
	+ \log D(r_{U},q_{\theta}(z_U|x_U))]
	\label{eq:adv_loss}
\end{equation}
where $z_L$, $z_U$ belong to the latent spaces for labeled, unlabeled data. 
The final training loss is 
${L}_{VAE}^{trans} + \lambda {L}_{VAE}^{adv}$.

The discriminator $D$ with ranking 
is learned to distinguish if latent space variable belongs to labeled pool. 
The loss is 
\begin{align}
{L}_{D} = 
	&-\mathbb{E}[\log D(r_{L},q_{\theta}(z_L|x_L))] \nonumber \\ 
	&-\mathbb{E}[ \log(1 - D(r_{U},q_{\theta}(z_U|x_U)))].
	\label{eq:disc_loss}
\end{align}
The smaller the output $D$ is, the more likely unlabeled sample is selected.  
The overall training pipeline 
is in Algorithm~\ref{algo-1} where
$\nabla_\theta$ denotes the gradient with respect to $\theta$.

After training TA-VAAL, the data points $(x^{\ast}_{1},...,x^{\ast}_{b})$ to be labeled at each stage are selected by
\begin{equation}
	(x^{\ast}_{1},...,x^{\ast}_{b}) = \mathop{\arg \min}_{(x_1,...,x_{b)} \subset X_U}D(R(x_U),q_{\theta}(z_U|x_U)).
	\label{eq:actlearn}
\end{equation}
The detail of selecting samples is described in supplemental.
A subset method, replacing $X_U$ in (\ref{eq:actlearn}) with
a random subset of $X_U$, was used to reduce outliers as suggested in~\cite{yoo2019learning}.

\section{Experimental Results}
\label{s:experiments}

\subsection{An illustrative example: binary classification}
\noindent
\textbf{Dataset.} The dataset with 2-dimensional features for binary classification
was generated using scikit-learns makemoons library~\cite{pedregosa2011scikit} as illustrated in Figure~\ref{fig:Toy}: 
The noise option was set to 0.2 and the
dataset size was 500 samples for one class and 50 samples for the other class, 
eventually constituting a dataset of size 550 
(imbalance ratio of class is $\times$10).

\noindent
\textbf{Implementation details.}
For the task learner, a 3-layer multi-layer perceptron (MLP) was used 
and Adam optimizer with learning rate 0.1 was used. 
For Ranker, a single layer perceptron was attached to the mid-layer of the task learner.
For VAE, a 2-layer MLP with ReLU was used for encoder and decoder, respectively, and
the discriminator comprised of a 2-layer MLP. For both the VAE and the discriminator, the Adam
optimizer with learning rate 0.01 was used. All epochs were set to 100 and active sampling was
performed starting from 20 random samples with 10 sample increment. 

\noindent
\textbf{Results.} 
Task-aware learning loss method tends to select difficult and informative samples that are all close to decision boundary, but are often clustered due to no information about global distribution even after performing a random subset method (left subfigure of Figure~\ref{fig:Toy}). 
In contrast, task-agnostic VAAL tends to select influential samples that are spread spatially, but that are often far from decision boundary due to no task related information (middle of Figure~\ref{fig:Toy}).
Our proposed TA-VAAL tends to select difficult (close to decision boundary) and influential (over the entire distribution) samples due to task-aware ranking information and data distribution-based VAAL, respectively (right of Figure~\ref{fig:Toy}). 

\begin{figure}[!b]
		\vspace{-1em}
	\centering	
	\begin{subfigure}{.9\linewidth}
		\centering
		\includegraphics[width=0.9\linewidth]{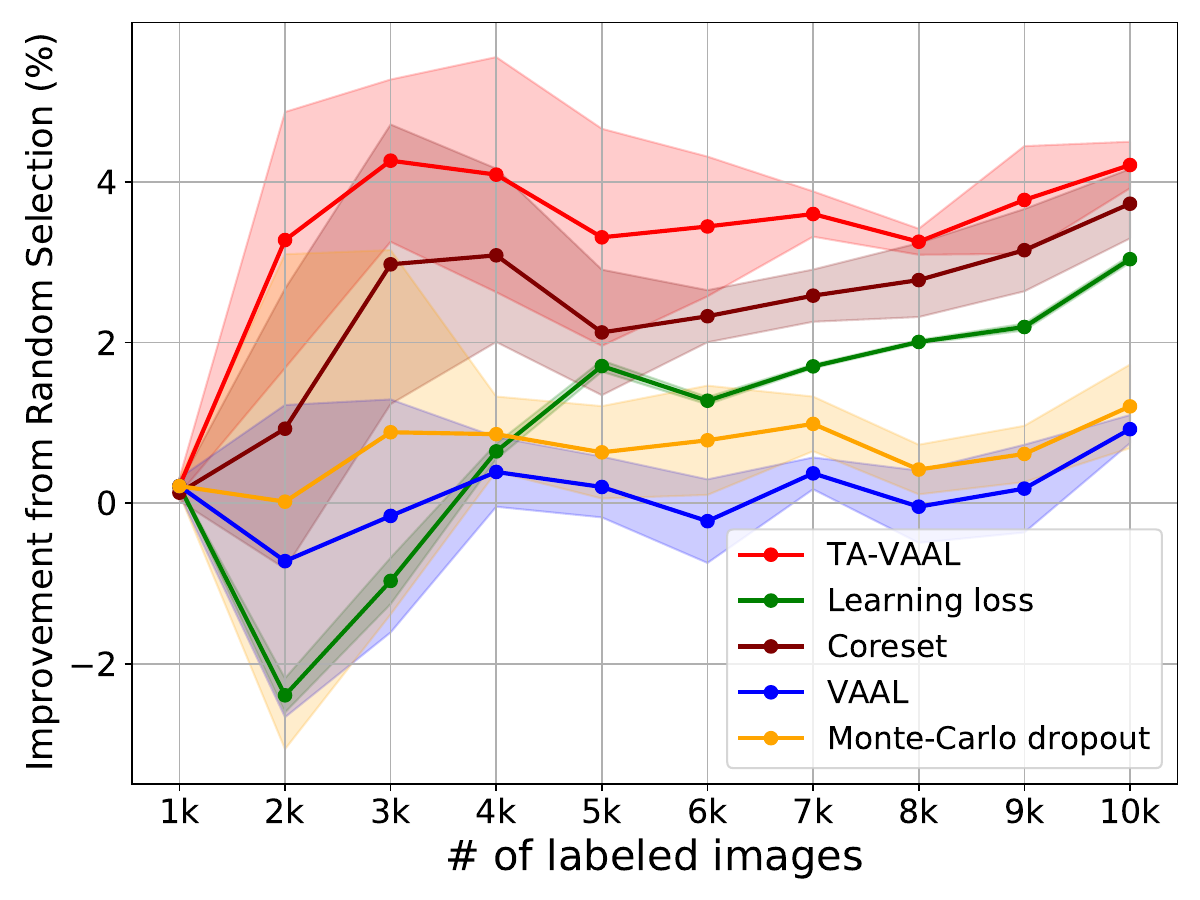}  
				\vspace{-.5em}
		\caption{CIFAR10}
		\label{fig:balance-first}
	\end{subfigure}
	\begin{subfigure}{.9\linewidth}
		\centering
		\includegraphics[width=0.9\linewidth]{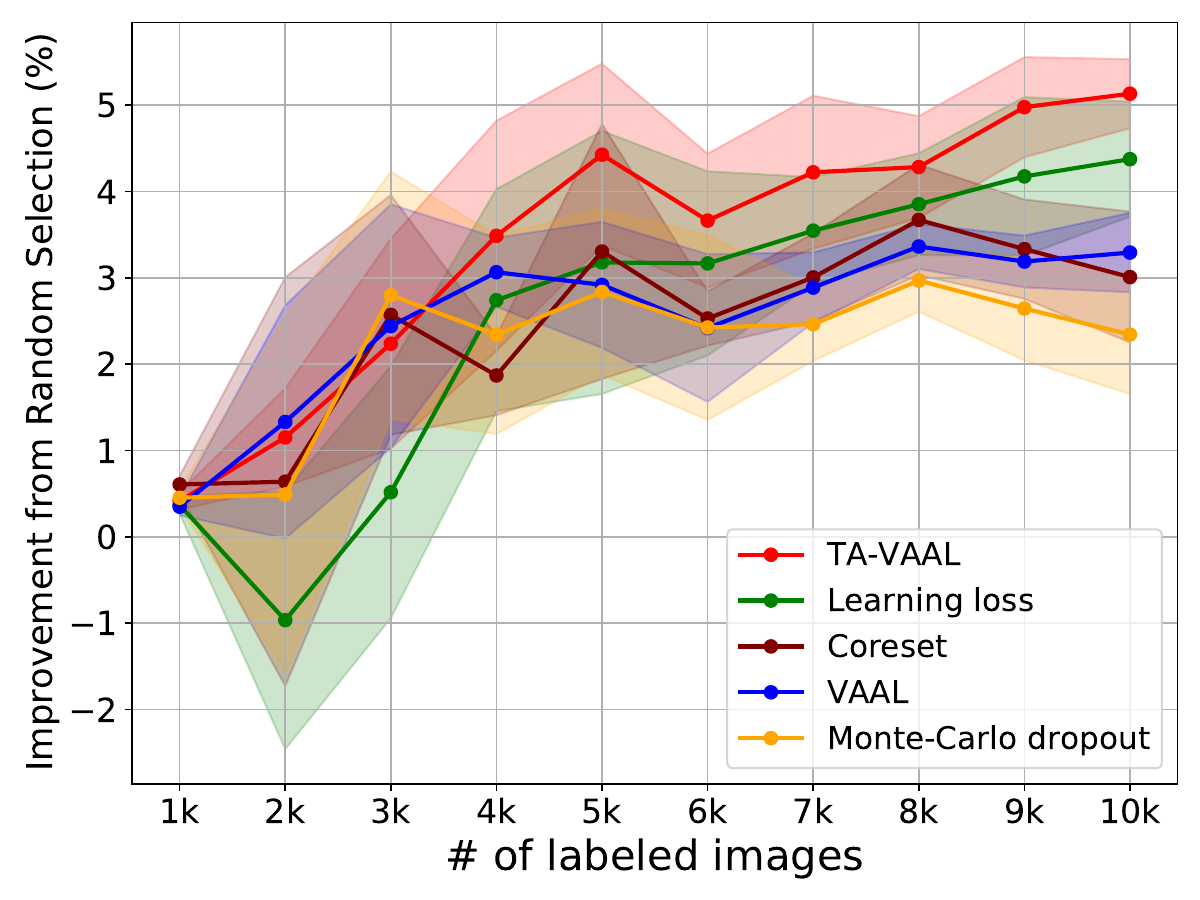}  
				\vspace{-.5em}
		\caption{CIFAR100}
		\label{fig:balance-second}
	\end{subfigure}
		\vspace{-1em}
	\caption{Mean accuracy improvements with standard deviation (shaded)
	of AL methods 
	from random sampling baseline over the number of labeled samples. 
	The absolute accuracy values are provided in the supplemental material. 
    Our TA-VAAL outperformed others on (balanced) CIFAR10 in all stages and on (balanced) CIFAR100 after a few stages.}
	\label{fig:balance}
\end{figure}

\subsection{Image classification on balanced datasets}
\noindent
\textbf{(Balanced) benchmark datasets.} We evaluated our proposed TA-VAAL method on various (balanced) benchmark datasets:
CIFAR10~\cite{krizhevsky2009cifar}, CIFAR100~\cite{krizhevsky2009learning} that
consist of 50,000 / 10,000 $32 \times 32$ images,
for training / testing with 10, 100 classes, respectively.
Each class includes the same number of images (5,000 / 1,000 images per class for CIFAR10,
500 / 100 images per class for CIFAR100).
The numbers of initial random samples were 1,000 / 1,000 with the query sizes 1,000, 1,000 at each stage on CIFAR10 / CIFAR100, respectively.
The results of the evaluations for our TA-VAAL on SVHN~\cite{netzer2011svhn} and Fashion-MNIST~\cite{xiao2017fashion} datasets are available in supplemental.
The subset method was used to avoid overlaps and to introduce diversity in samples: the subset size was set to 10 times larger than the query size. 
\noindent
\textbf{Implementation details.}
For training, 
$32\times32$ random crop from $36 \times 36$ zero-padded images,
normalization with mean and standard deviation of training set, 
and horizontal flip / flop augmentation were used. 
ResNet18~\cite{he2016deep} was used for all task learners and stochastic gradient descent (SGD) 
was used with momentum 0.9 and weight decay 0.005.
Learning rate was 0.1 for the first 160 epochs and then 0.01 for the last 40 epochs.
For VAE, a modified Wasserstein auto-encoder~\cite{tolstikhin2017wasserstein} 
for taking ranking information was used and the discriminator was constructed as a 5-layer MLP. 
For both the VAE and the discriminator, 
Adam optimizer~\cite{kingma2014adam} with learning rate $5 \times 10^{-4}$ was used. 
Mini-batch size was 128 and the total epochs were 200 for all datasets.

\noindent
\textbf{Results.} 
Six AL methods were evaluated: random sampling (baseline), Monte-Carlo dropout~\cite{gal2017deep}, Core-set~\cite{sener2017active}, Learning loss~\cite{yoo2019learning}, VAAL~\cite{sinha2019variational} and
our TA-VAAL. 
Figure~\ref{fig:balance} presents the number of labeled images (active samples) versus the mean (line) and standard deviation (shaded region) for accuracy improvements from the baseline with 5 trials.

In Figure~\ref{fig:balance-first} for CIFAR10, learning loss method yielded even lower accuracy than baseline at early stages possibly due to insufficient labeled samples to capture the uncertainty of model and yielded good performance at later stages once sufficient labeled data was used to train learning loss. VAAL achieved better performance than learning loss possibly due to massive unlabeled data.
Our proposed TA-VAAL outperformed other state-of-the-art methods at all stages.

In Figure~\ref{fig:balance-second} for CIFAR100, all active learning methods outperformed baseline (random sampling) in most stages. Learning loss method exhibited similar tendency (low performance at early stages, then high performance at later stages) on both CIFAR10 and CIFAR100. After 3k labeled samples, our TA-VAAL outperformed all compared state-of-the-art active learning methods substantially.

\noindent
\textbf{Discussions.}
1) Core-set yielded comparable performance to our TA-VAAL on CIFAR10. 
However, Core-set is computationally demanding as compared to ours since 
core-set required 7.5 times more selection time per sample than our TA-VAAL.
2) We used much smaller initial data size / budget (1,000/1,000) than the original VAAL setting (5,000/2,500) on CIFAR10~\cite{sinha2019variational} and
VAAL yielded similar performance as random sampling for all cases in our setting (see supplemental for detail).  
3) The performances of learning loss and ours yielded slightly higher or lower mean accuracies at the first stage 
due to additional LPM attached to the task learner. 
We performed additional study 
to show that this additional loss is not the most important factor for the overall performance improvements of our proposed method (see supplemental).

\begin{figure}[!t]
	\centering	
	\begin{subfigure}{.9\linewidth}
	\centering
	\includegraphics[width=0.9\linewidth]{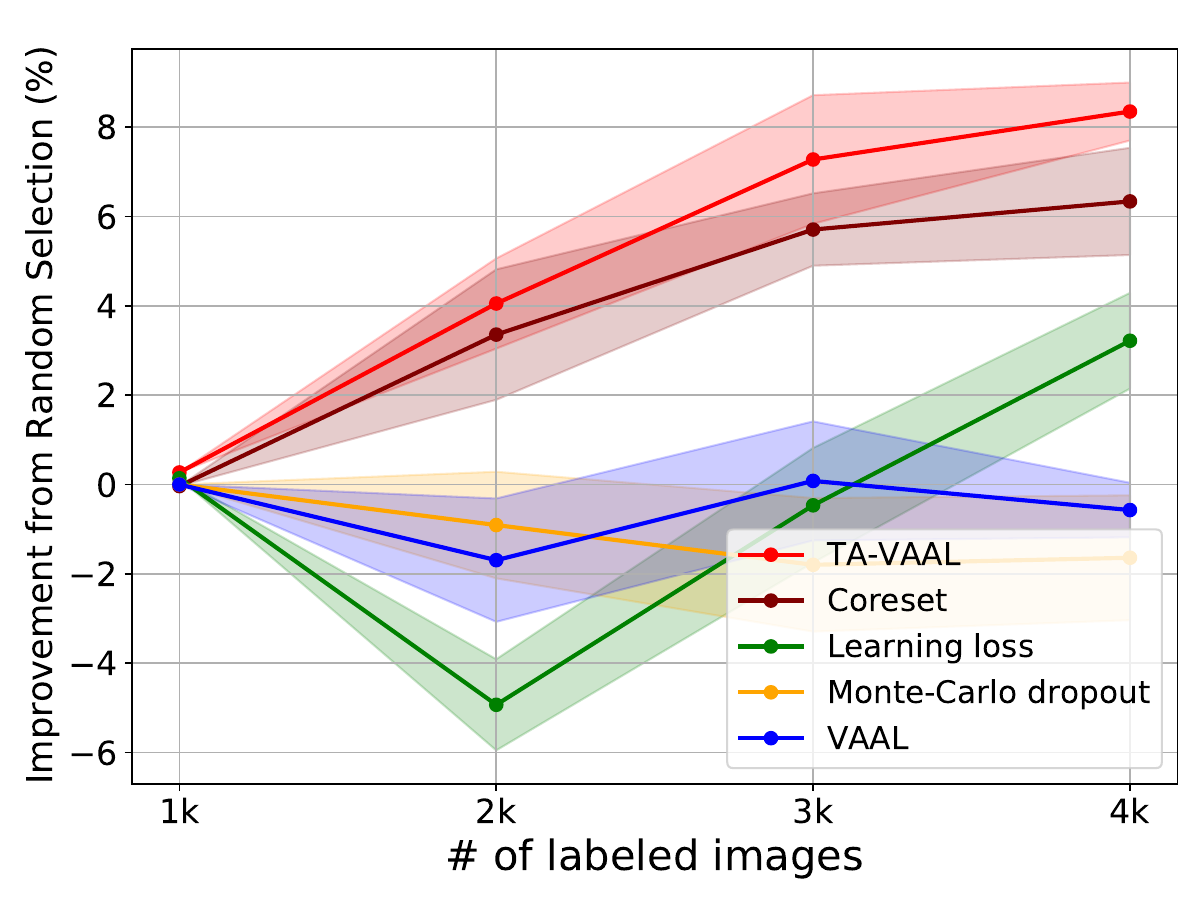}
					\vspace{-.5em}
	\caption{Modified CIFAR10 with imbalance ratio $\times$100}
	\label{fig:imbalance-first}
	\end{subfigure}
	\begin{subfigure}{.9\linewidth}
	\centering
	\includegraphics[width=0.9\linewidth]{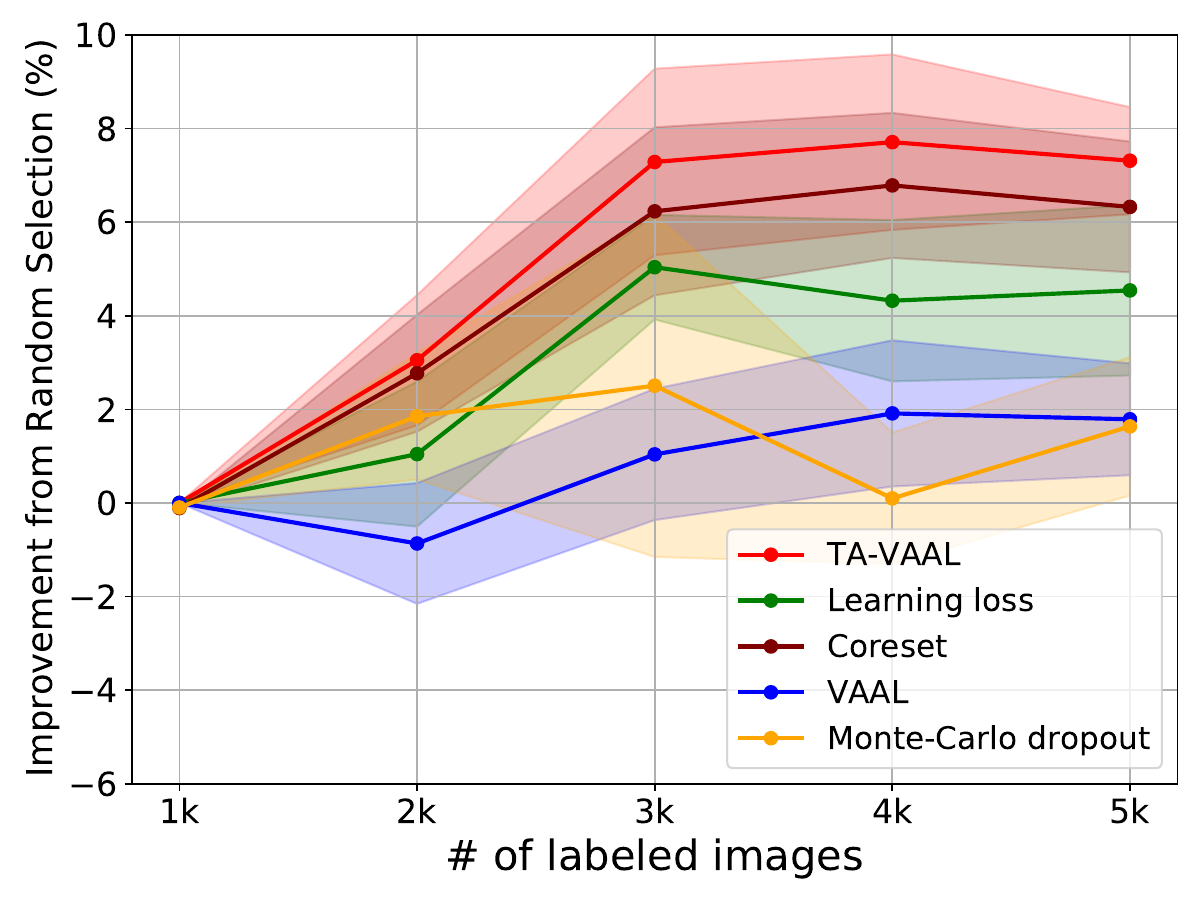}
					\vspace{-.5em}
	\caption{Modified CIFAR10 with imbalance ratio $\times$10}
	\label{fig:imbalance-second}
	\end{subfigure}
	\begin{subfigure}{.9\linewidth}
		\centering
		\includegraphics[width=0.9\linewidth]{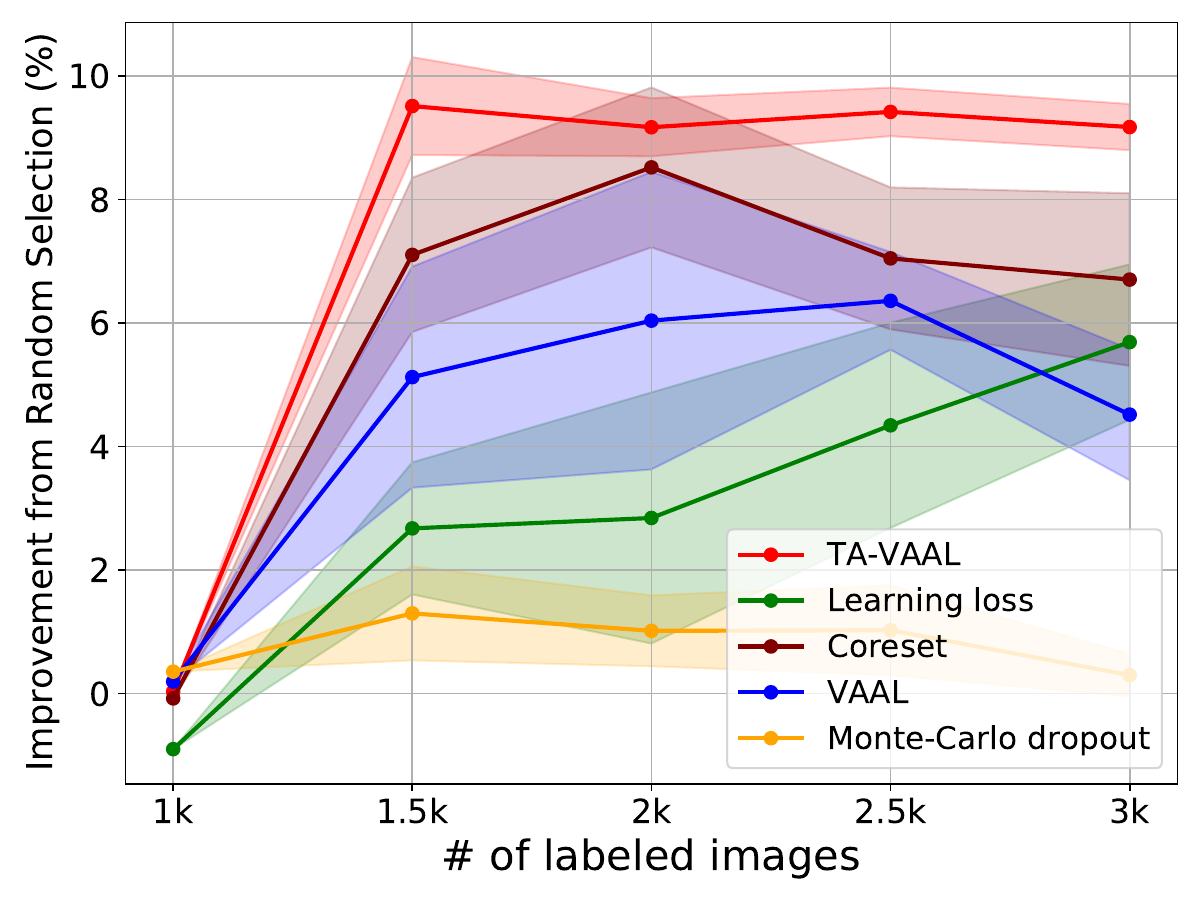}  
				\vspace{-.5em}
		\caption{Caltech101}
		\label{fig:Caltech}
	\end{subfigure}
			\vspace{-1em}
	\caption{Mean accuracy improvements with standard deviation (shaded) of AL methods from random sampling baseline over the number of labeled samples on imbalanced datasets.
    Our TA-VAAL outperformed others on modified CIFAR10 with different imbalance ratios and
    Caltech101 in all stages.}
	\label{fig:imbalance}
				\vspace{-1em}
\end{figure}

\subsection{Image classification on imbalanced datasets}
\noindent
\textbf{Datasets.} We performed experiments on imbalanced datasets whose sizes are different for classes.
Modified CIFAR10's were constructed by randomly reducing the number of samples that were associated with the first 5 classes. Imbalance ratio was defined as the ratio of the number of samples for the first 5 classes to the number of samples for the last 5 classes. Imbalance ratios of $10$ and $100$ were used.

Further evaluation was performed on Caltech101~\cite{fei2006one} that
consists of 9,144 images with 
about 300 $\times$200 and 101 categories with imbalanced labels (40 - 800 images per class, mostly 50).
We set 8,125 images for training and 1,019 images for testing. 
Initially, 1,000 images were randomly selected and AL budget was 500 images per stage. 

\noindent
\textbf{Implementation detail.} For modified CIFAR10, we used the same implementation for CIFAR10.
For Caltech101, we performed random horizontal flips for data augmentation and resized images to 224$\times$224 for training. ResNet18 was used as the task learner and SGD was used with learning rate of $0.01$.
Modified Wasserstein autoencoder and 5-layer MLP were used for conditional VAE and discriminator, respectively. 
Adam optimizer with learning rate 1 $\times10^{-4}$ was used with minibatch size 16 and 200 epochs. The details on hyper-parameters are described in supplement material.

\noindent
\textbf{Results.}
Figures~\ref{fig:imbalance-first} and~\ref{fig:imbalance-second} illustrate the mean and standard deviation accuracy improvements from random sampling baseline (5 trials) over the number of labeled images on two modified CIFAR10 with imbalance ratios of $\times$10 (less imbalanced) and $\times$100 (more imbalanced).
Note that reduced data size in imbalanced datasets limited the maximum number of experiments up to 4k-5k labeled samples.
Our proposed TA-VAAL outperformed all other state-of-the-art methods over all stages with more improvement margins for more imbalanced dataset ($\times$100 imbalance ratio).
Note that even though the final dataset sizes were 4k and 5k, there were some classes with 50 total images per class for imbalance ratio $\times$100. In this challenging case, our TA-VAAL still yielded improvements over other methods including random sampling baseline. 
See supplemental for absolute accuracy over the number of labeled images for all methods.

Figure~\ref{fig:Caltech} presents the number of labeled images vs. the mean and standard deviation for accuracy improvements from random sampling (5 trials) on (naturally imbalanced) Caltech101. 
Our TA-VAAL outperformed other state-of-the-art methods over all stages substantially. These results show the capability of our TA-VAAL in a more realistic setting with more classes with label imbalance and larger images.

\begin{figure}[!b]
				\vspace{-1em}
	\begin{subfigure}{0.9\linewidth}
		\centering
		\includegraphics[width=0.9\linewidth]{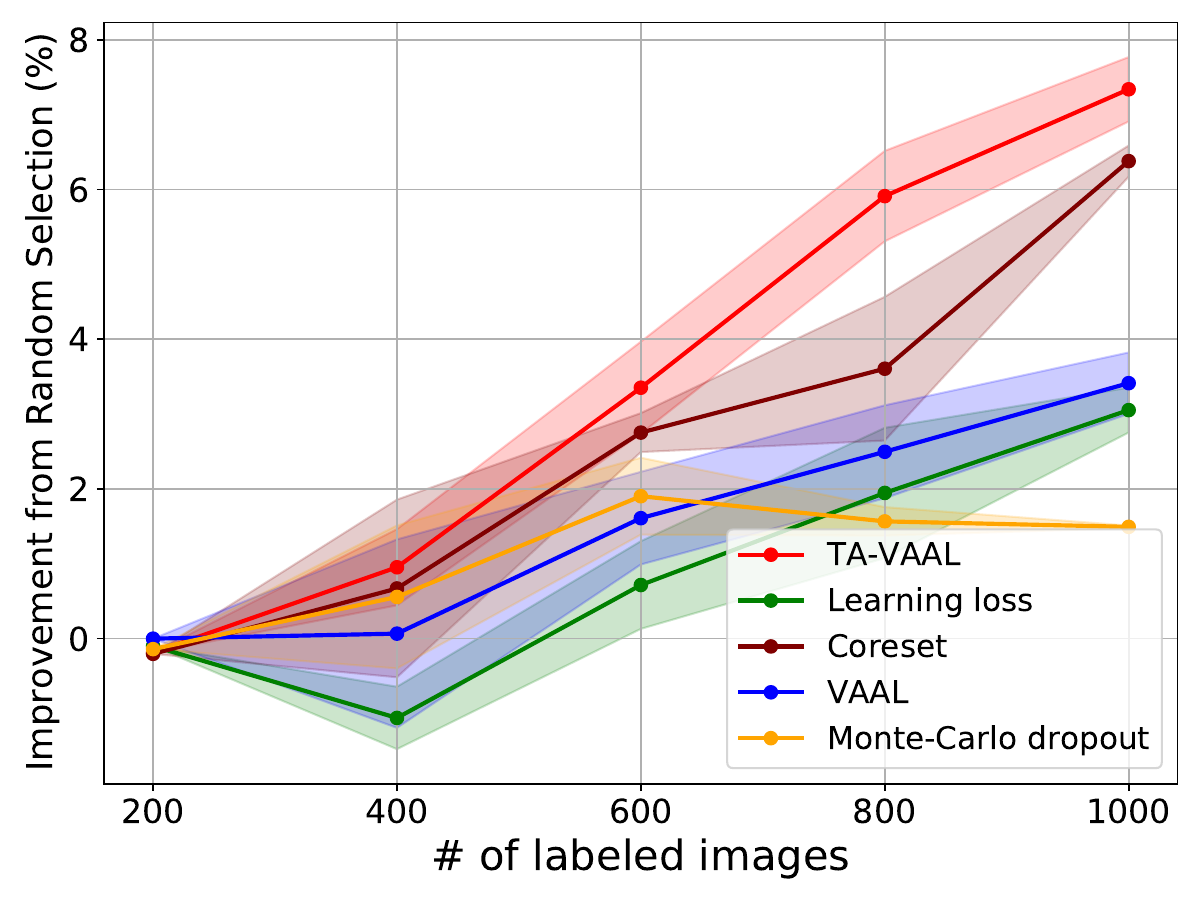}  
	\end{subfigure}
				\vspace{-1em}
	\caption{Relative accuracy improvements from random selection for semantic segmentation on Cityscape dataset.}
	\label{fig:Cityscape}
\end{figure}

\subsection{Semantic segmentation on Cityscapes}

\noindent
\textbf{Dataset.} 
AL was performed for semantic segmentation on Cityscapes~\cite{cordts2016cityscapes}, a large-scale video dataset of street scenes,
including 3,475 frames with instance segmentation annotations.
Following~\cite{yu2017dilated}, we converted labels into 19 classes. 
Initial label pool size was 200 with budget size 200 per stage. 

\noindent
\textbf{Implementation detail.} For training, we performed random horizontal flips for data augmentation similar to classification tasks. 
We adopt DRN~\cite{yu2017dilated} as task learner for image segmentation and SGD was used with learning rate $1 \times 10^{-3}$.  
Modified Wasserstein autoencoder and 5-layer MLP were used for conditional VAE and discriminator, respectively. 
Adam with learning rate 1 $\times 10^{-4}$ was used with mini-batch size 4 and total epoch 100.
See supplemental for details.

\noindent
\textbf{Results.} Six AL methods were evaluated including random sampling (baseline), Monte-Carlo dropout~\cite{gal2017deep}, Core-set~\cite{sener2017active}, Learning loss~\cite{yoo2019learning}, VAAL~\cite{sinha2019variational} and
our TA-VAAL. Figure~\ref{fig:Cityscape} shows the number of labeled images versus the mean IoU (Intersection Over Union) of 3 trials. We observed that our TA-VAAL outperformed all compared methods
at all sampling stages. VAAL yielded better performance than random sampling and learning loss at all stages, but our TA-VAAL outperformed other methods with substantial margins at all sampling stages.
This demonstrated the benefits from the ranking loss information of task learner to select the most informative samples from unlabeled pool.  

\section{Empirical Analyses}

\begin{figure}[!b]
			\vspace{-1em}
	\centering
	\includegraphics[width=0.9\linewidth]{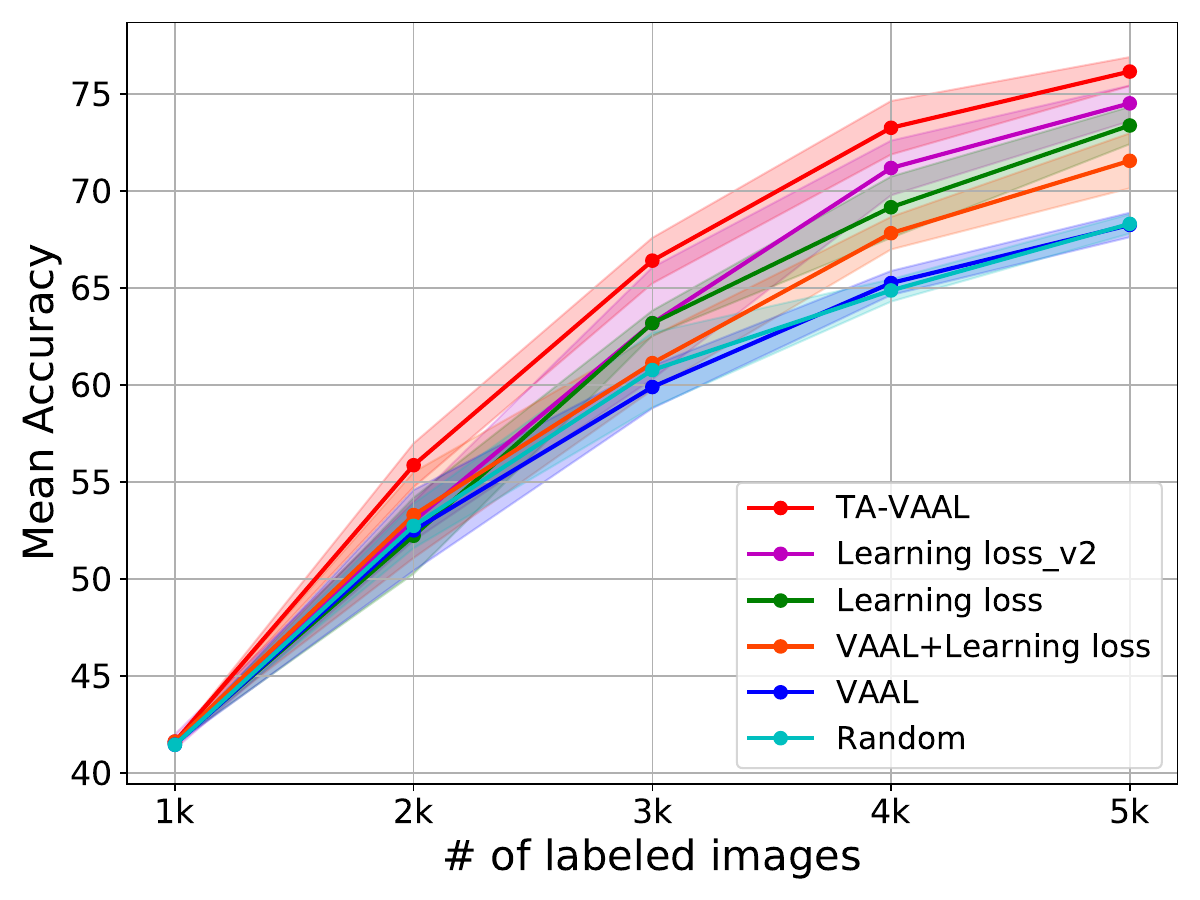}
					\vspace{-1em}
	\caption{The results of ablation study by selectively removing core components (modified CIFAR10 with imbalanced ratio $\times$10): Learning loss\_v2 is ours without VAAL. VAAL+learning loss is ours with the original learning loss.}
	\label{fig:ablation}
\end{figure}

\subsection{Ablation studies}
\label{sec:ablation}

Figure~\ref{fig:ablation} shows the performance results of our proposed methods with and without proposed components / structures along with
other state-of-the-art methods. The means and standard deviations of 5 trials were reported.
Firstly, learning loss method with proposed ranking loss (\ref{eq:ranking_loss}), called learning loss\_v2, 
yielded substantially higher performances at later AL
stages and comparable performances at early stages to the original learning loss. 
Thus, it seems that using our proposed loss (\ref{eq:ranking_loss}) for accurate loss ranking prediction seems advantageous over using the original LPM loss for accurate loss prediction.
Another study is to incorporate ranking information into VAAL by using the original learning loss architecture, rather than our proposed Ranker (\ref{eq:ranking_loss}).
This combination of VAAL+learning loss still yielded substantially better performances than VAAL over all stages. However,
that was not able to yield better performance than the original learning loss method at later stages.

\begin{figure}[!t]
	\centering
	\begin{subfigure}{1\linewidth}
		\centering
		\includegraphics[width=0.9\linewidth]{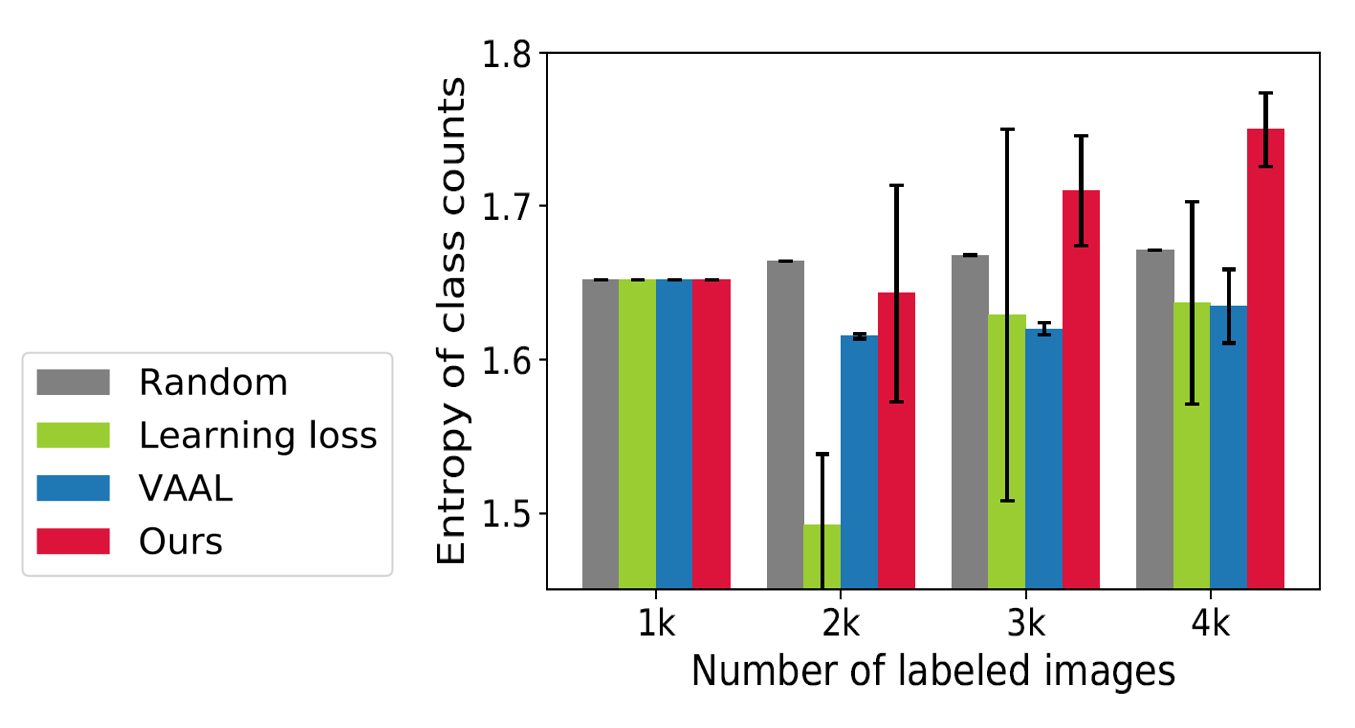} 
			\vspace{-.5em}
		\caption{} \label{fig:entropy}
	\end{subfigure}
	\hspace{.05\textwidth}	
	\begin{subfigure}{0.9\linewidth}
		\centering
		\includegraphics[width=0.8\linewidth]{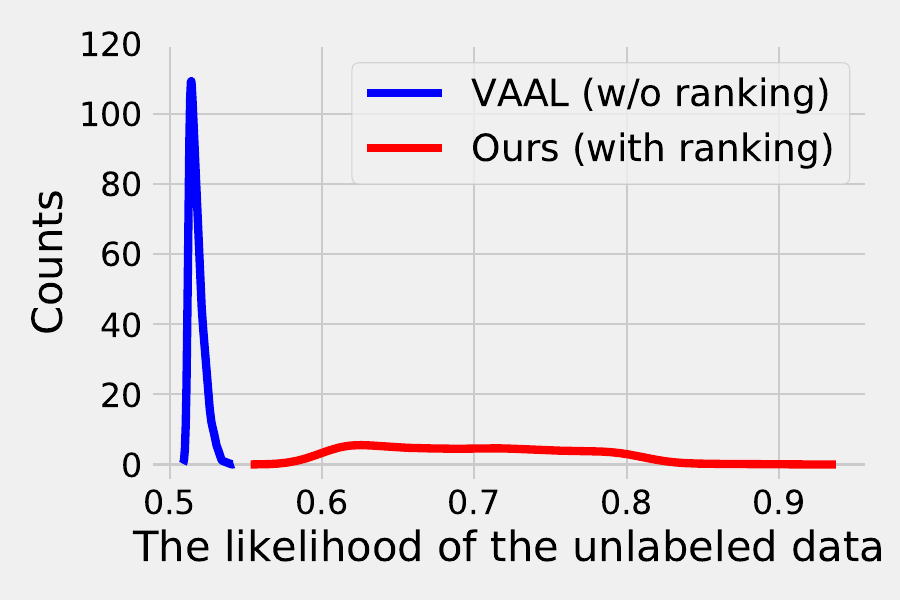}
			\vspace{-.5em}
		\caption{} \label{fig:discrinimator}	
	\end{subfigure}		
	\vspace{-1em}
	\caption{For the modified CIFAR10 with imbalance ratio $\times 100$,
		(a) Bar graphs of number of labeled images vs. data class count entropy.
		(b) Likelihood of unlabeled data vs. number of samples at the last stage.
	}
		\vspace{-1em}
\end{figure}

\begin{figure}[!t]
	\centering
	\begin{subfigure}{0.9\linewidth}
		\centering
		\includegraphics[width=0.9\linewidth]{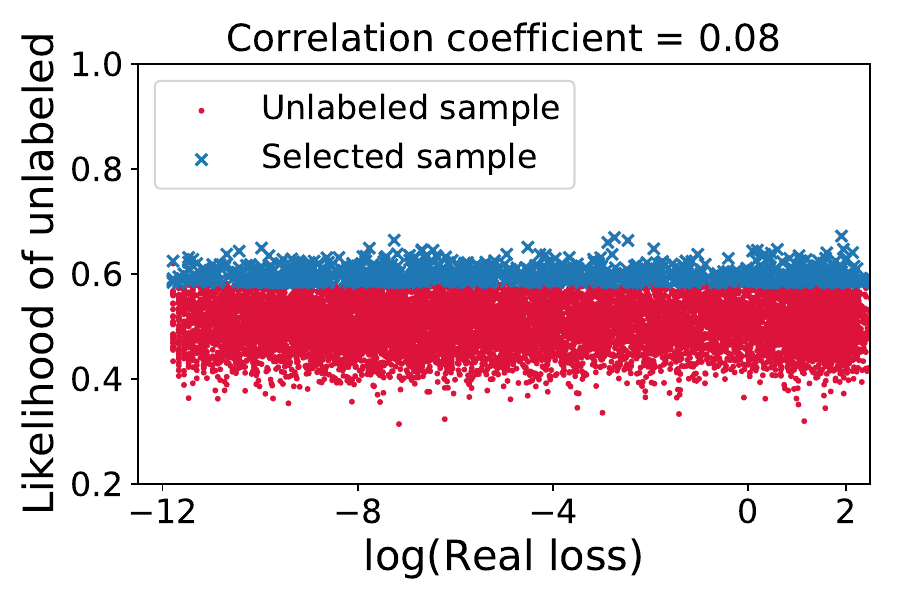}
			\vspace{-.5em}
		\caption{VAAL} \label{fig:analysis_vaal}
	\end{subfigure}
	\hspace{.05\textwidth}	
	\begin{subfigure}{0.9\linewidth}
		\centering
		\includegraphics[width=0.9\linewidth]{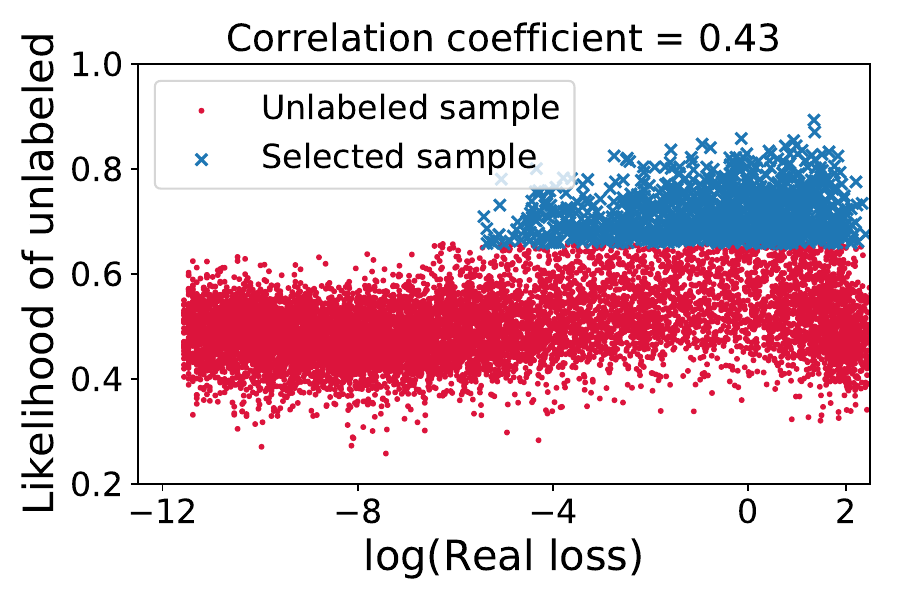}
			\vspace{-.5em}
		\caption{Our TA-VAAL} \label{fig:analysis_ours}
	\end{subfigure}
		\vspace{-1em}
	\caption{Relationships between real loss (task model uncertainty) and likelihood of data remaining unlabeled (task-agnostic data distribution) in (a) VAAL and (b) our TA-VAAL. 
		We use the model from the last AL stage 
		on imbalanced CIFAR-10.
		While task-agnostic VAAL selected samples with a wide range of real loss values, our TA-VAAL chose samples with relatively high real loss values.} 
	\label{fig:analysis}
		\vspace{-1em}
\end{figure}

\subsection{On selected samples of active learning}

Figure~\ref{fig:entropy} shows the bar graphs for the number of labeled images (selected samples) versus the entropies of the number of selected samples over 10 classes (class counts).  
The higher the entropy is, the more uniform samples over classes are selected. Figure~\ref{fig:entropy} shows that our proposed method selected samples with high data class count entropy on a severely imbalanced dataset.
These results can provide insights to explain the performance results in Figure~\ref{fig:imbalance-first}. For example, learning loss method yielded substantially low performance at 2k stage in Figure~\ref{fig:imbalance-first} due to its data selection with low data count entropy over classes at that stage as illustrated in Figure~\ref{fig:entropy}. 
This is possibly due to limited number of data for certain classes in the case of imbalance ratio $\times$100 so that
task learner in learning loss method was not well-trained. 
However, our TA-VAAL was able to select good samples at the same stage due to
the structure from VAAL to exploit overall data distribution 
so that good performance and high data class count entropy were able to be achieved.

Figure~\ref{fig:discrinimator} shows the likelihood of unlabeled data from the discriminator $D$
to select data points at the last stage. 
VAAL that takes latent space values as discriminator input yielded concentrated count distribution of the likelihood (from the output of $D$) at the last stage so that active learning selection became almost random, while our proposed method that takes latent space values along with ranking information for $D$ yielded a wide range of likelihood distribution so that sample selection was more reliable and yielded good performances as illustrated in Figures~\ref{fig:entropy} and~\ref{fig:imbalance-first},~\ref{fig:imbalance-second}.

Figure~\ref{fig:analysis} shows the graphs for (true) real loss (representing task-aware model uncertainty) vs. likelihood of data remaining unlabeled (representing task-agnostic data distribution) for VAAL and our proposed TA-VAAL. Note that both VAAL and our TA-VAAL methods select data points that have the highest estimated likelihood of unlabeled data. 
While task-agnostic VAAL selected samples with a wide range of real losses as illustrated in Figure~\ref{fig:analysis_vaal}, our proposed TA-VAAL was able to choose samples with relatively high real loss values thanks to the reshaped latent space by ranking information on real losses as shown in Figure~\ref{fig:analysis_ours}.
Thus, TA-VAAL seems to result in higher performances in various tasks over VAAL as in Figure~\ref{fig:balance}.

\section{Conclusion}

We proposed TA-VAAL, a novel AL framework that simultaneously takes advantage of both 
task-agnostic data distribution-based AL and task-aware model uncertainty-based approach
that exploits any generic task learner (with or without latent space).
Our TA-VAAL exploits VAAL that considered data distribution of both label and unlabeled pools by incorporating LPM and RankCGAN concepts into VAAL by relaxing loss prediction with a ranker for ranking loss information.
We demonstrate that our TA-VAAL outperforms state-of-the-art AL methods on various classification benchmark datasets such as CIFAR-10, CIFAR-100 and Caltech-101 for balanced and imbalanced cases and on Cityscapes semantic segmentation dataset. Our in-depth analyses also confirm that our TA-VAAL effectively takes advantage of both task-aware and task-agnostic AL approaches.

\pagebreak

\section{Supplementary Section}

In this supplemental document, we present:

\quad $\bullet$ brief summaries of  
active data sampling strategy of our Task-Aware Variational Adversarial Active Learning (TA-VAAL), 
accompanied by the corresponding algorithm descriptions (Sec. \ref{algo});

\quad $\bullet$ performance comparisons of 
``learning loss'' methods with the original loss 
and our modified ranking loss (called learning loss\_v2 in the main paper) (Sec. \ref{compare});

\quad $\bullet$ absolute average accuracies and mIOU of active learning results in the main paper. (Sec. \ref{acc}); 

\quad $\bullet$ active learning results on CIFAR100 with a larger budget size that is consistent with the work of VAAL. (Sec. \ref{large});

\quad $\bullet$ details of hyperparameters for training in TA-VAAL (Sec. \ref{hyper});

\quad $\bullet$ example images that were selected at each stage of different active learning algorithms (Sec. \ref{example});

\subsection{Algorithm for active sampling strategy}
\label{algo}
Algorithm~\ref{algo-2} details the sample selection strategy at each stage of the TA-VAAL training
process. 
The goal of each sample selection strategy is collecting $b$ number of samples from unlabeled data pool $X_U$ to update the labeled pool.
First, we predicted the ranking of loss from Ranker, denoted by $R(;\theta_{R})$, for unlabeled data pool to obtain task-aware (model-uncertainty) information: 
\begin{equation*}
	r_U \leftarrow R(x_U;\theta_{R}), \quad \forall x_U \in X_U;
\end{equation*}
This value will be normalized so that the absolute loss value will not be preserved while the relative ranking information will be preserved before embedded into the latent space of VAE.
Second, we got the latent space values from the encoder of VAE, denoted by $q_{\theta}(\cdot)$, for unlabeled pool:
\begin{equation*}
	z_U \leftarrow q_{\theta}(z_U|x_U), \quad \forall x_U \in X_U;
\end{equation*}
Finally, we selected the data points $(x_{1}^{\ast} ,..,x_{b}^{\ast})$ by the following operation:
\begin{equation*}
(x^{\ast}_{1},...,x^{\ast}_{b}) = \operatorname*{argmin}_{(x_1,...,x_{b)} \subset X_U}D(R(x_U),q_{\theta}(z_U|x_U));
\end{equation*}
Note that the smaller the output of discriminator $D$, the more likely its latent space belongs to unlabeled pool. The main idea of our approach is that rather than relying only on latent space that represents the probability from unlabeled pool, our proposed method utilized task-aware information from Ranker that represents the score that task learner predicts with low confidence to select \emph{influential} and \emph{difficult} data points. 
\begin{algorithm}[ht]
	\caption{Active sampling in TA-VAAL}
	\SetKwInOut{Input}{Input}
	\SetKwInOut{Output}{Output}
	\SetKwInput{kwset}{Set}
	\Input{budget size $b$ and unlabeled data pool $X_U$}
	\Output{acquisition data samples $(x_{1}^{\ast} ,..,x_{b}^{\ast})$}
	\enskip Predict ranking of loss for unlabeled data pool:
	\begin{equation*}
	r_U \leftarrow R(x_U;\theta_{R}), \quad \forall x_U \in X_U;
	\end{equation*}
	Get latent space values for unlabeled data pool:
	\begin{equation*}
	z_U \leftarrow q_{\theta}(z_U|x_U), \quad \forall x_U \in X_U;
	\end{equation*}
	Choose the data points $(x_{1}^{\ast} ,..,x_{b}^{\ast})$ by the following operation:
	\begin{equation*}
	(x^{\ast}_{1},...,x^{\ast}_{b}) = \operatorname*{argmin}_{(x_1,...,x_{b)} \subset X_U}D(R(x_U),q_{\theta}(z_U|x_U));
	\label{eq:sampling}
	\end{equation*}
	\label{algo-2}
\end{algorithm}	

\subsection{Loss comparisons for loss prediction module}
\label{compare}
\begin{figure}
	\centering
	\includegraphics[width=0.45\textwidth]{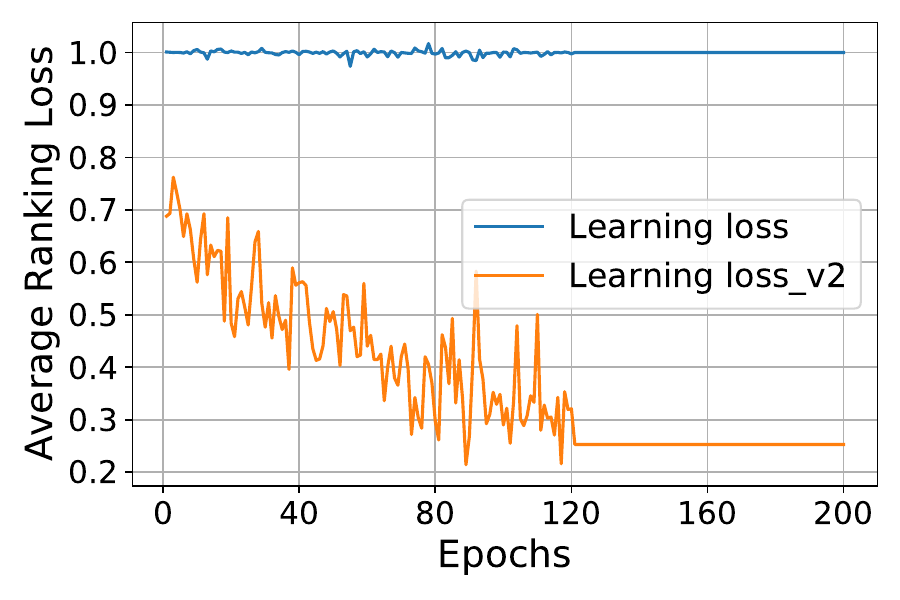}  
	\caption{Losses for the loss prediction module over the number of epochs in learning loss and learning loss$\_v2$.}
	\label{fig:rankingloss}
\end{figure}

Figure~\ref{fig:rankingloss} shows the graph of average loss versus the number of epochs 
for learning loss methods with the original learning loss and our modified learning loss, called learning loss\_v2.  
After the 120th epoch, we did not back propagate the gradient of Ranker so that the fluctuation of loss value is stopped as implemented in the original learning loss method.
We observe that the loss for the loss prediction module of learning loss (blue line) is not minimized possibly due to fixed $\epsilon = 1$ to emphasize more on the exact loss predictions, less on the relative rankings of them. However, the ranking loss of our modified learning loss (orange line) can be minimized as iteration continues possibly due to the relaxed condition for predicting loss values so that training seems to be more stable than the original learning loss. 

\subsection{Absolute accuracies}
\label{acc}
In the main paper, we showed the improvements of accuracy from the random sampling baseline over the number of labeled dataset. In this section, we showed the absolute accuracy curves over the number of labeled dataset in Figures~\ref{acc-1} and \ref{acc-2}.

\subsection{Another setting on initial size \& budget}
\label{large}

\begin{figure}[t]
	\centering
	\includegraphics[width=0.45\textwidth]{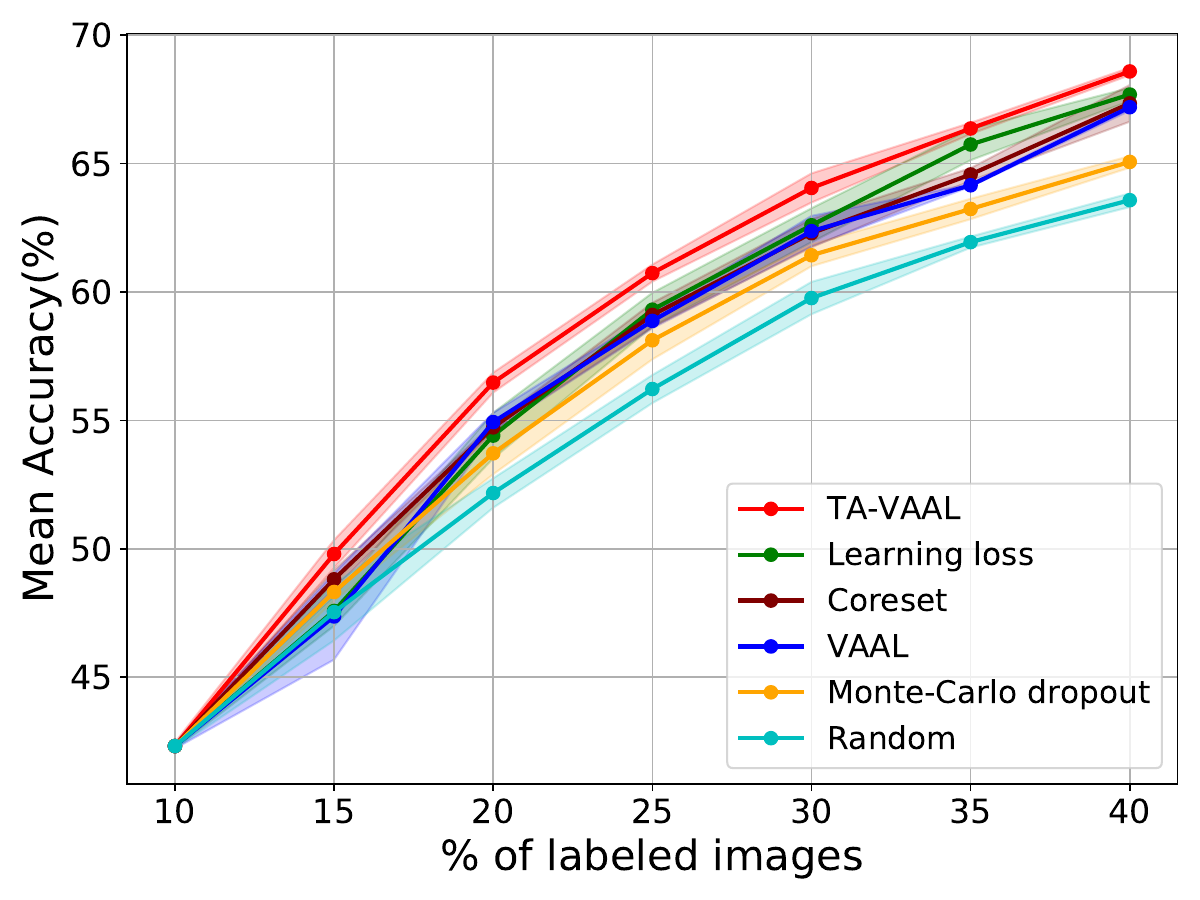}  
	\caption{Active learning results with a large budget size (5000 / 2500) on CIFAR100 dataset.}
	\label{fig:CIFARLARGE}
\end{figure}

\begin{figure*}[ht]	
	\centering
	\begin{subfigure}{.47\linewidth}
		\centering
		\includegraphics[width=1\linewidth]{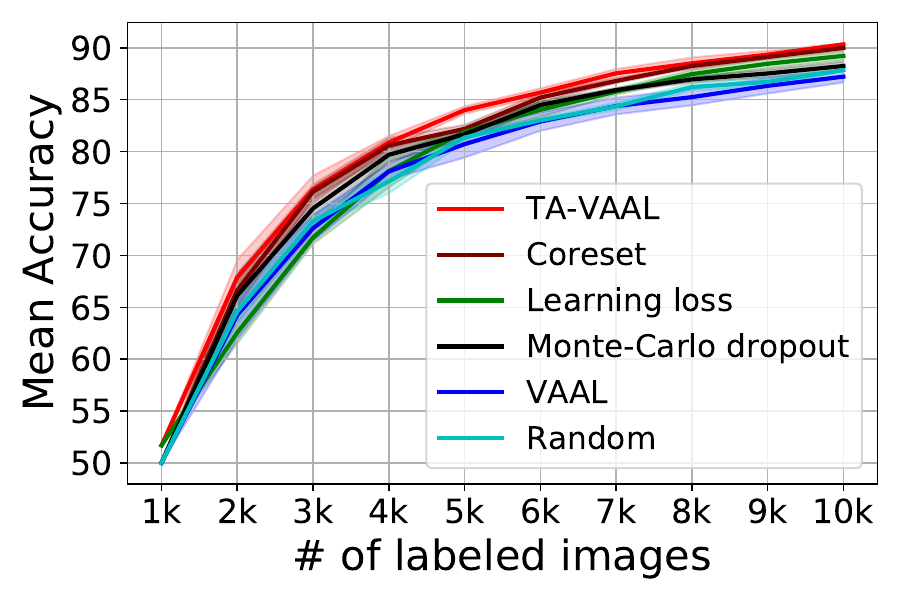}  
		\caption{CIFAR10}
		\label{fig:cifar10}
	\end{subfigure}
	\begin{subfigure}{.47\linewidth}
		\centering
		\includegraphics[width=1\linewidth]{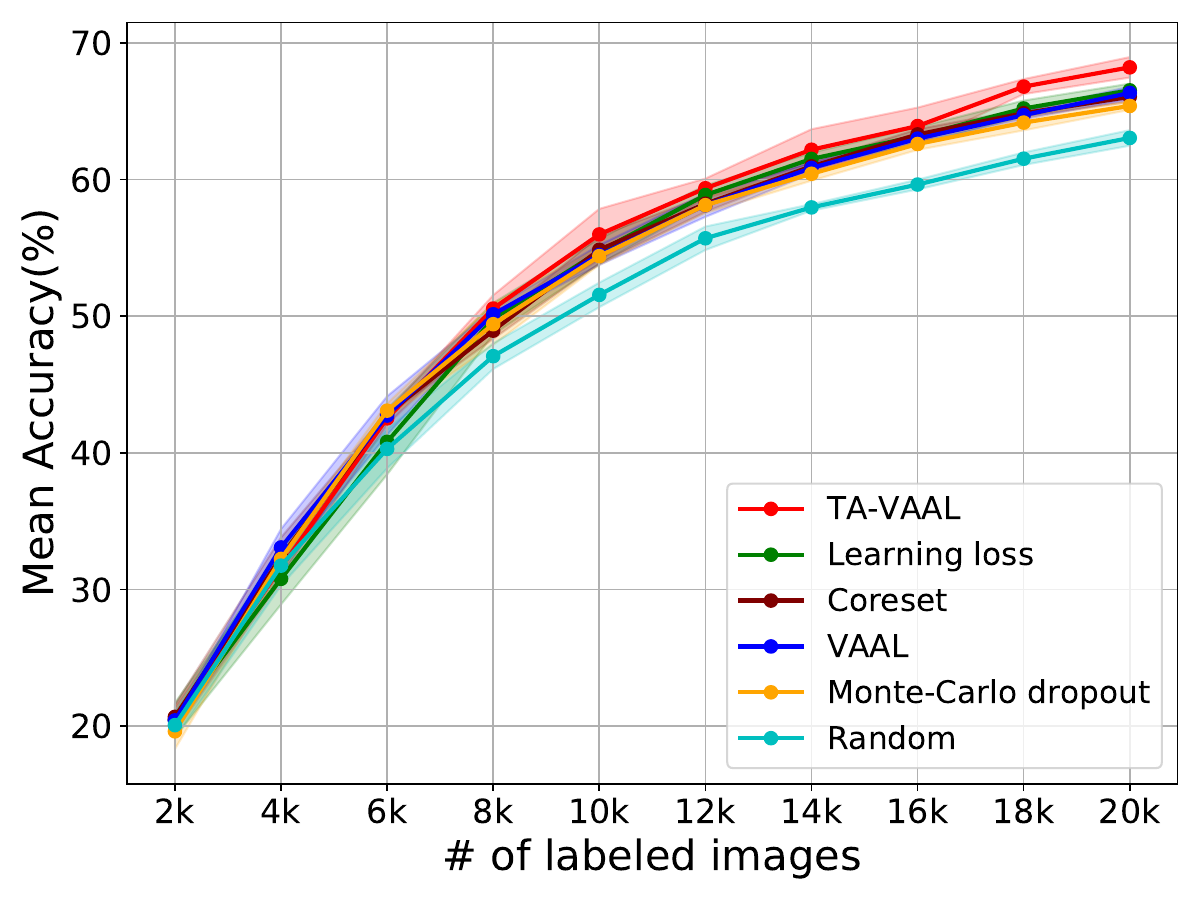}  
		\caption{CIFAR100}
		\label{fig:cifar100}
	\end{subfigure}
	\caption{
	Mean accuracy curves with standard deviation (shaded) of active learning methods over the number of labeled samples on (a) CIFAR10 and (b) CIFAR100.} 
	\label{acc-1}
\end{figure*}

\begin{figure*}[ht]	
	\centering
	\begin{subfigure}{.47\textwidth}
		\centering
		\includegraphics[width=1\linewidth]{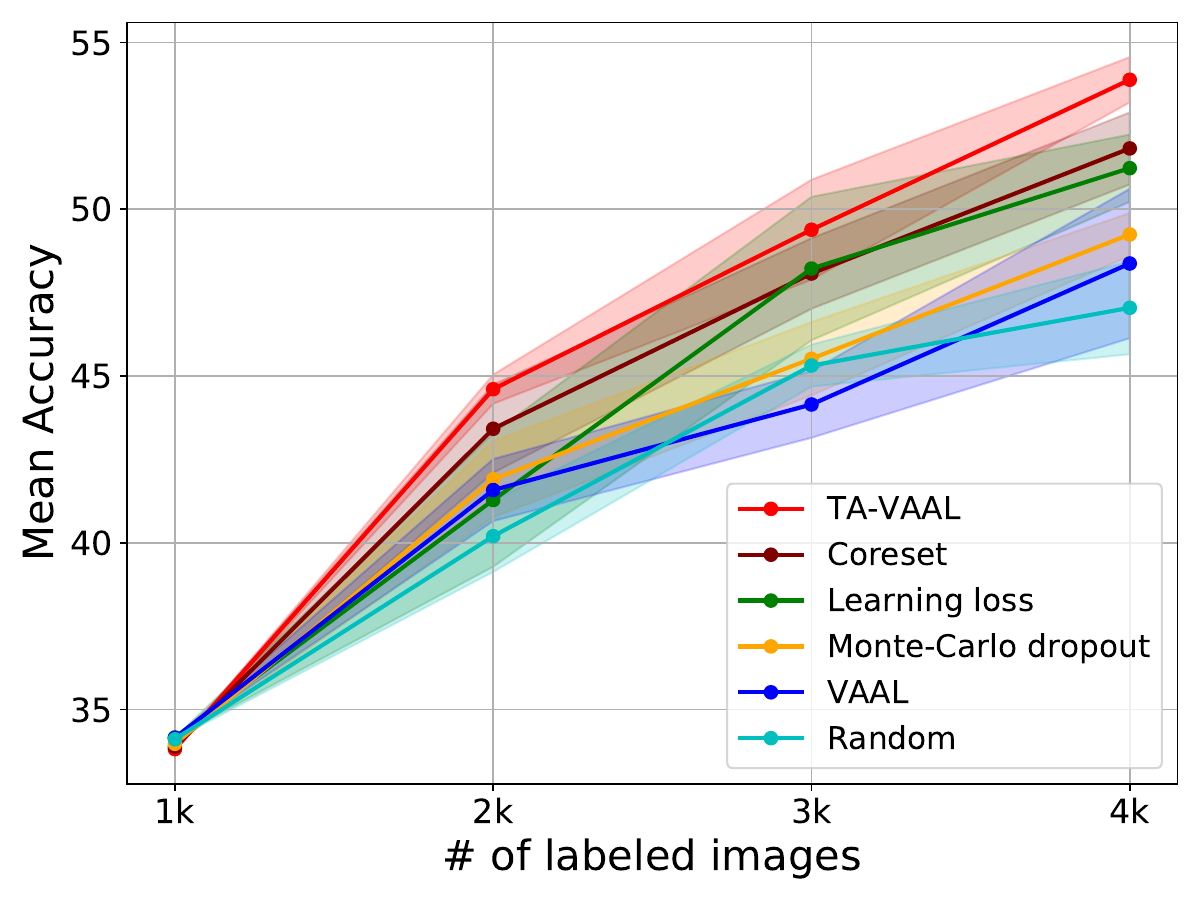}  
		\caption{Imbalance$\times$100}
		\label{fig:cifarx100}
	\end{subfigure}
	\begin{subfigure}{.47\textwidth}
		\centering
		\centering
		\includegraphics[width=1\linewidth]{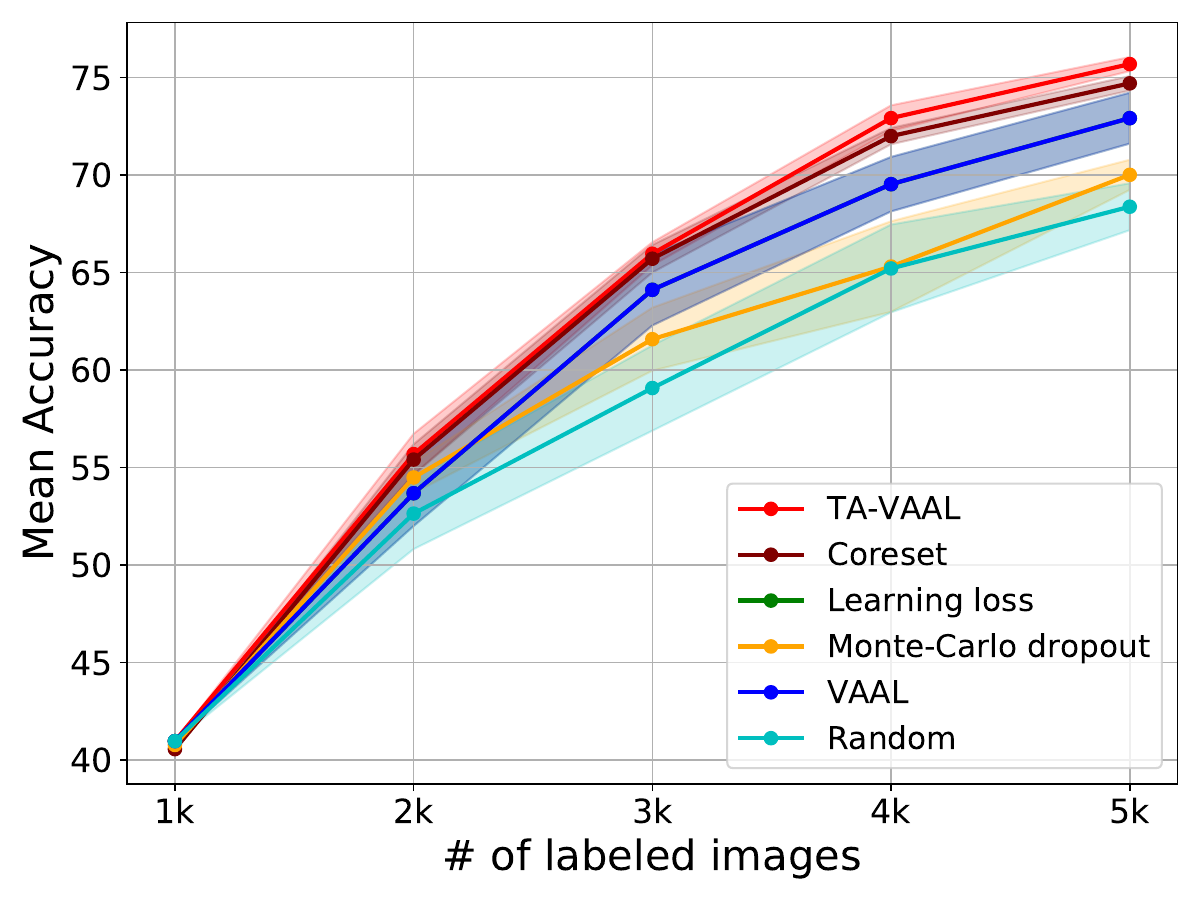}  
		\caption{Imbalance$\times$10}
		\label{fig:cifarx10}
	\end{subfigure}
	\begin{subfigure}{.47\textwidth}
		\centering
		\includegraphics[width=1\linewidth]{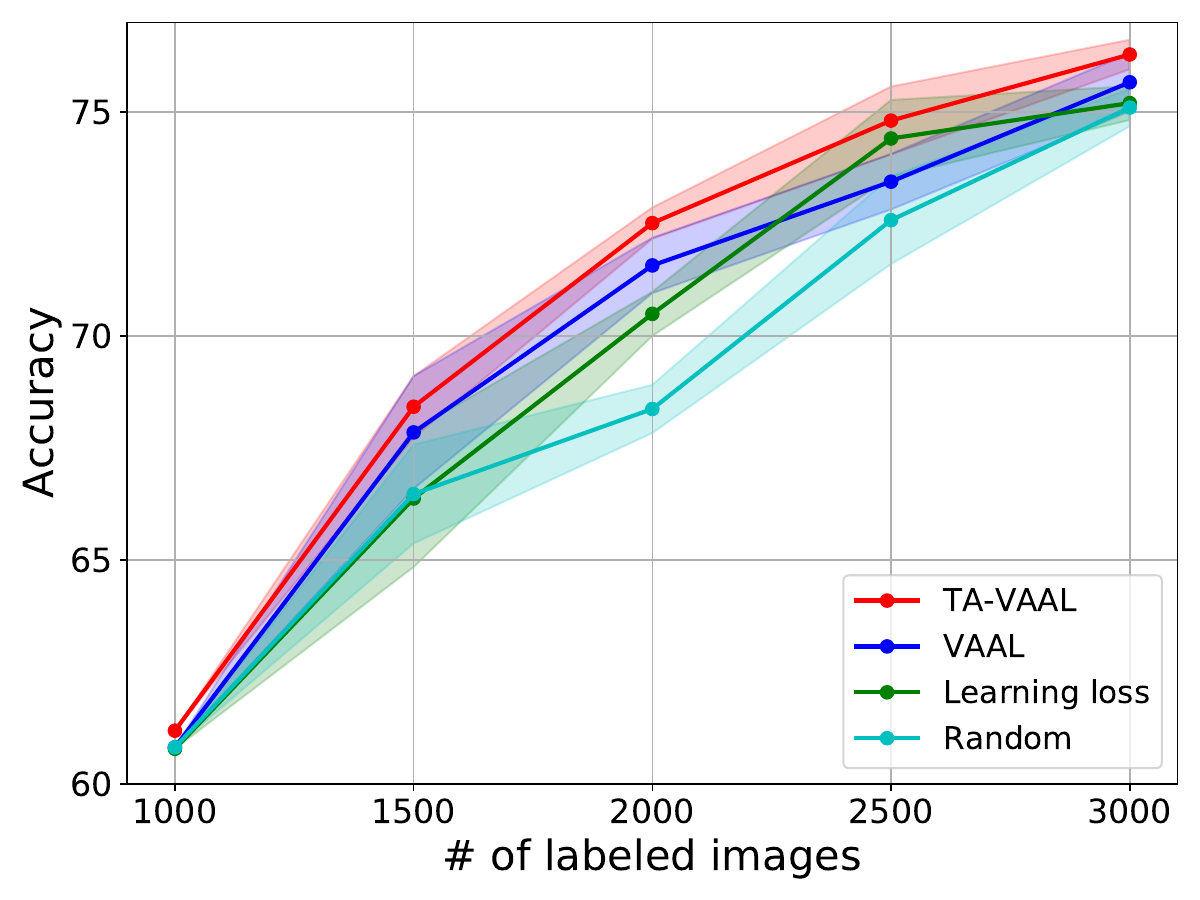}  
		\caption{Caltech101}
		\label{fig:caltech101}
	\end{subfigure}
	\begin{subfigure}{.47\textwidth}
		\centering
		\centering
		\includegraphics[width=1\linewidth]{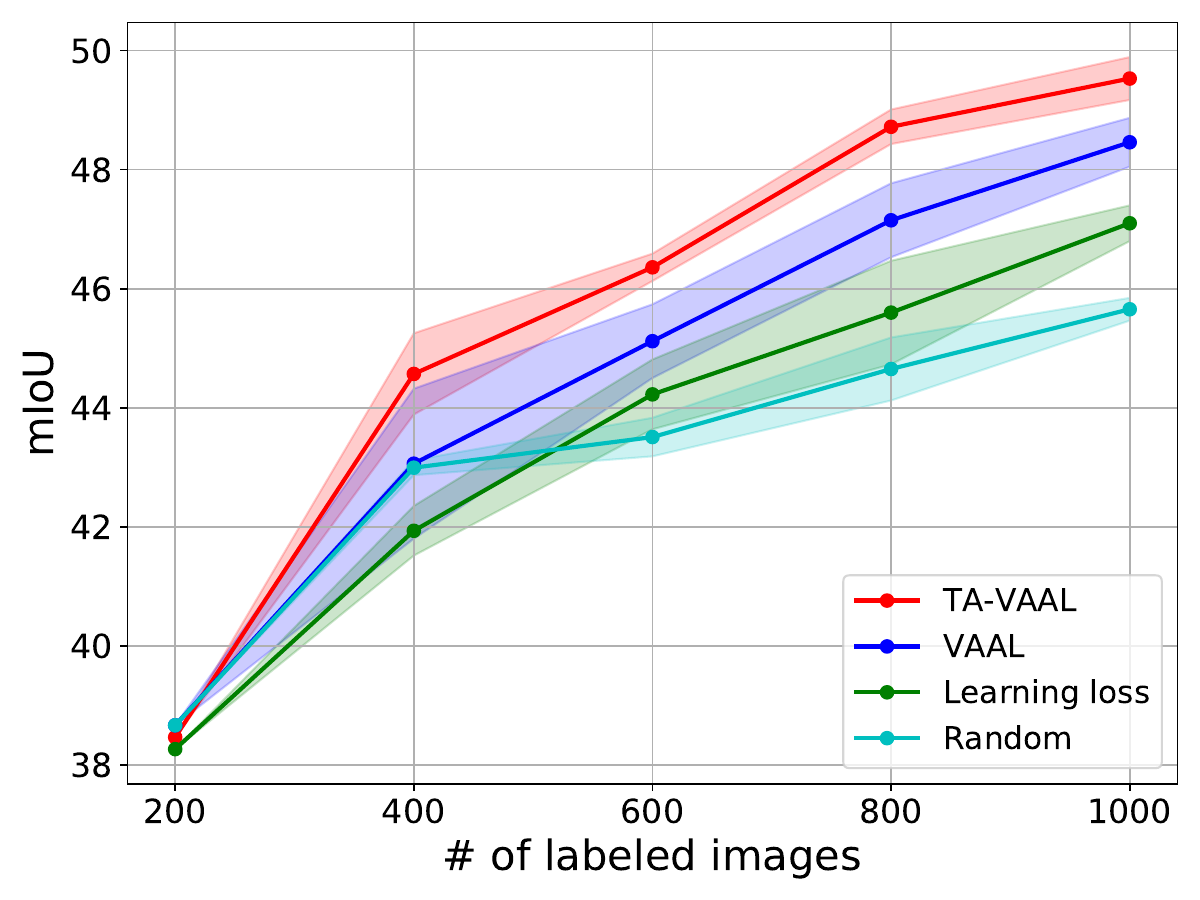}  
		\caption{Cityscape}
		\label{fig:cityscape}
	\end{subfigure}
	\caption{Mean accuracy or Mean intersection over union (IOU) curves with standard deviation (shaded) of active learning methods over the number of labeled samples for image classifications on (a) modified CIFAR10 with imbalanced $\times$100, (b) modified CIFAR10 with imbalanced $\times$10, (c) Caltech101 datasets and for semantic segmentation on (d) Cityscape dataset.} 
	\label{acc-2}
\end{figure*}

In the main paper, our experiments used smaller number of labels (1000 / 1000) or (2000 / 2000) that followed and extended the settings of the work of learning loss~\cite{yoo2019learning} than the reported VAAL experiments on CIFAR10, CIFAR100 (5000 / 2500)~\cite{sinha2019variational}. We argue that our setting with smaller initial size and budget would be more beneficial for active learning than larger initial size and budget. However, for those who are familiar with the work of VAAL~\cite{sinha2019variational}, we briefly validate our implementations with the same active learning setting as VAAL for CIFAR100 (or 5000 / 2500 for initial / budget sizes) and the results are presented in Figure~\ref{fig:CIFARLARGE}. Note that our proposed method still outperformed other state-of-the-art methods in all stages for the same setting in VAAL.

\begin{table*}[t!]
	\vspace{-10pt}
	\caption{Hyperparameters used in TA-VAAL. $d$ is the latent space dimension of VAE. $\zeta_1$, $\zeta_2$, and $\zeta_3$ are learning rates of task learner T, VAE, and discriminator $D$, respectively. $\eta$ is a scaling parameter for total loss of task learner in Eq. (2). $\lambda$ are regularization parameters for transductive and adversarial losses of VAE. $\beta$ is a Lagrangian parameter in Eq. (3). ``Initial'' represents that the size of labeled data pool at initial stage and ``budget'' indicates that the size of samples which to be selected at each stage. 
	Zero padding was used. Large images were cropped considering the trade-off between training speed and GPU resources.}
	\label{tabe:hyper}
	\begin{center}	
		\begin{small}
			\begin{tabular}{c|c|c|c|c|c|c|c|c|c|c|c}
				\toprule
				Dataset & $d$ & $\zeta_1$&$\zeta_2$&$\zeta_3$&$\eta$& $\lambda$&$\beta$&batch size& epochs&Inital / budget&image size \\
				\midrule
				CIFAR10     & 64 & 1$\times 10^{-1}$ & 5$\times 10^{-4}$ & 5$\times 10^{-4}$& 1 & 1 &1&128&200&1000 / 1000 &32 $\times$ 32 (padded)\\
				CIFAR100    & 64 & 1$\times 10^{-1}$ & 5$\times 10^{-4}$ & 5$\times 10^{-4}$& 1 & 1&1&128&200&2000 / 2000 &32 $\times$ 32 (padded)\\		
				Caltech101  & 128& 1$\times 10^{-2}$ & 1$\times 10^{-4}$ & 1$\times 10^{-4}$& 0.2 & 15 &1&16&200& 1000 / 500& 224 $\times$ 224 (cropped) \\   
				Cityscapes  & 128 & 1$\times 10^{-3}$ & 1$\times 10^{-4}$ & 1$\times 10^{-4}$& 0.1 & 25 &1&4&100& 200 / 200 & 412 $\times$ 412 (cropped)  \\   
				\bottomrule
			\end{tabular}
		\end{small}
	\end{center}
\end{table*}

\subsection{Detail on hyperparameters}
\label{hyper}
Table~\ref{tabe:hyper} shows the hyperparameters for training our proposed method for different datasets.
We set these hyperparameters based on VAAL settings and tuned these through a grid search.  

\subsection{Example sampled images}
\label{example}
\begin{figure*}[ht]	
	\begin{subfigure}{.33\textwidth}
		\centering
		\includegraphics[width=0.9\linewidth]{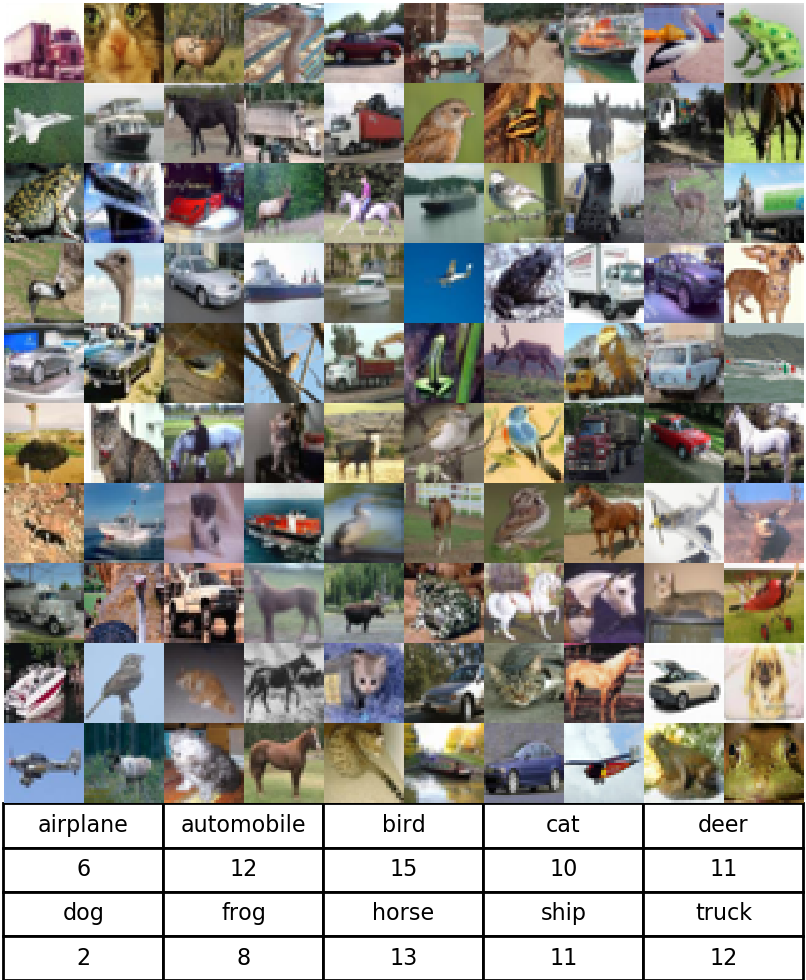}  
		\caption{Learningloss - stage 1}
		\label{fig:Learning-2k}
	\end{subfigure}
	\begin{subfigure}{.33\textwidth}
		\centering
		\includegraphics[width=0.9\linewidth]{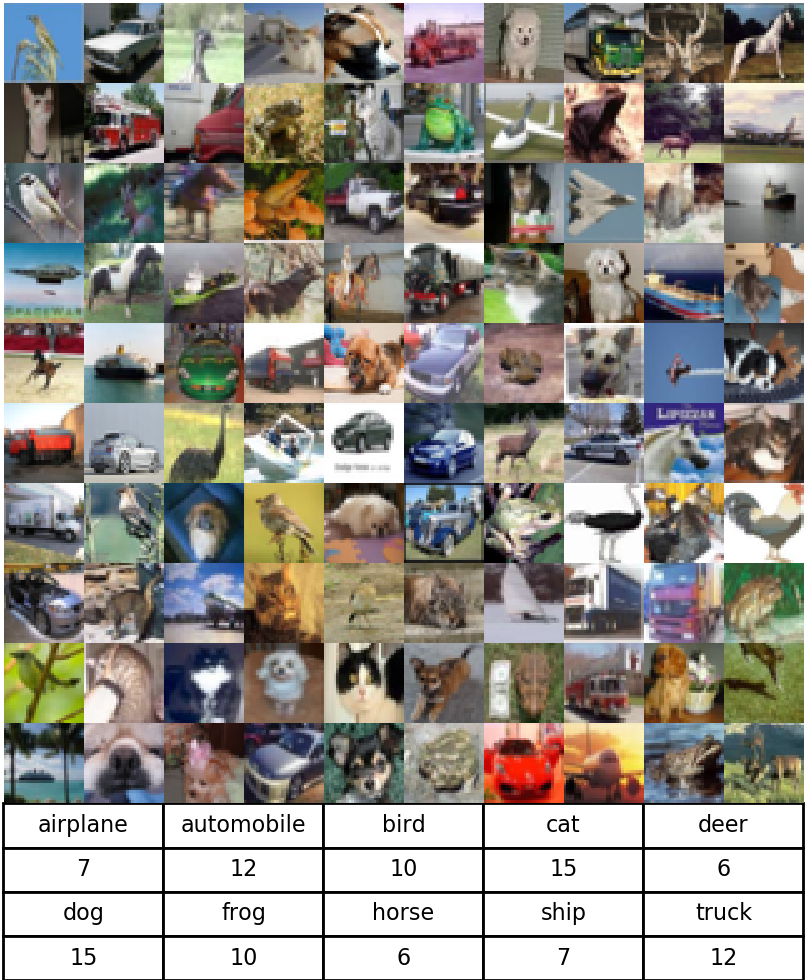}  
		\caption{Learningloss - stage 2}
		\label{fig:Learning-3k}
	\end{subfigure}
	\begin{subfigure}{.33\textwidth}
		\centering
		\includegraphics[width=0.9\linewidth]{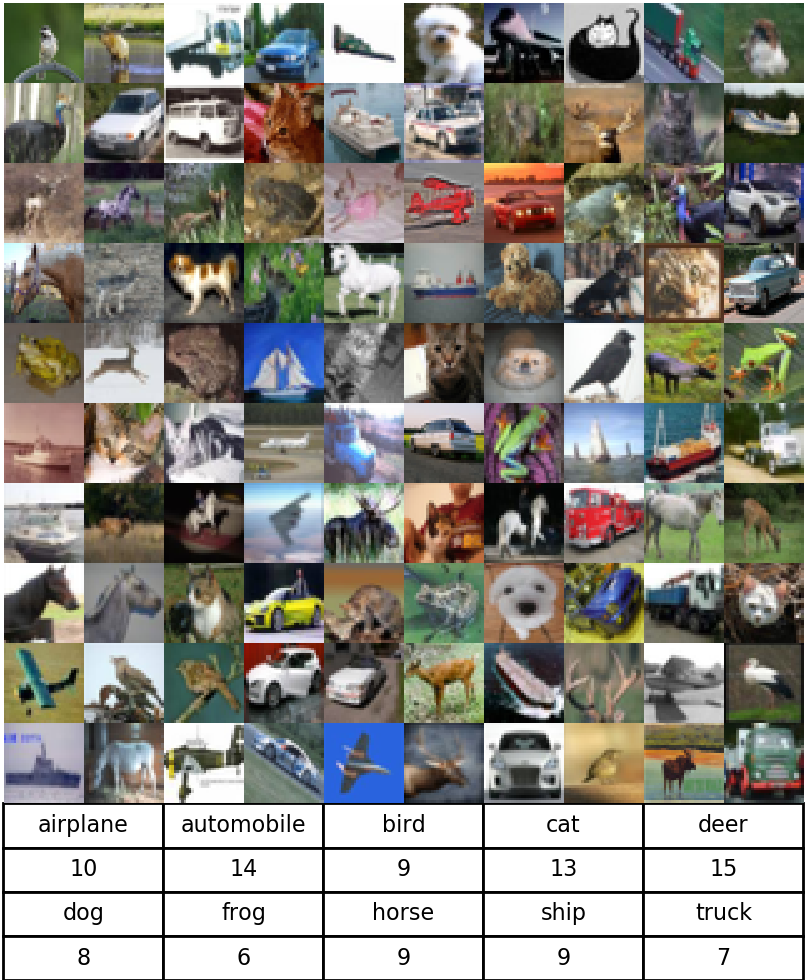}  
		\caption{Learningloss - stage 3}
		\label{fig:Learning-4k}
	\end{subfigure}
	\begin{subfigure}{.33\textwidth}
		\centering
		\includegraphics[width=0.9\linewidth]{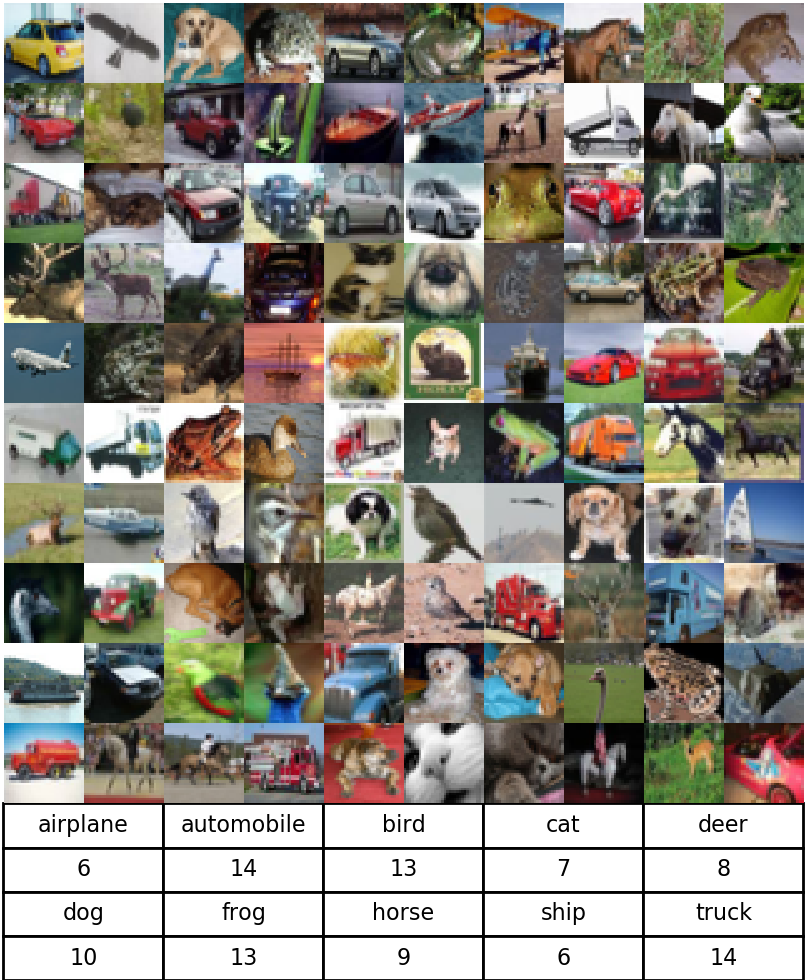}  
		\caption{VAAL - stage 1}
		\label{fig:VAA-2k}
	\end{subfigure}
	\begin{subfigure}{.33\textwidth}
		\centering
		\includegraphics[width=0.9\linewidth]{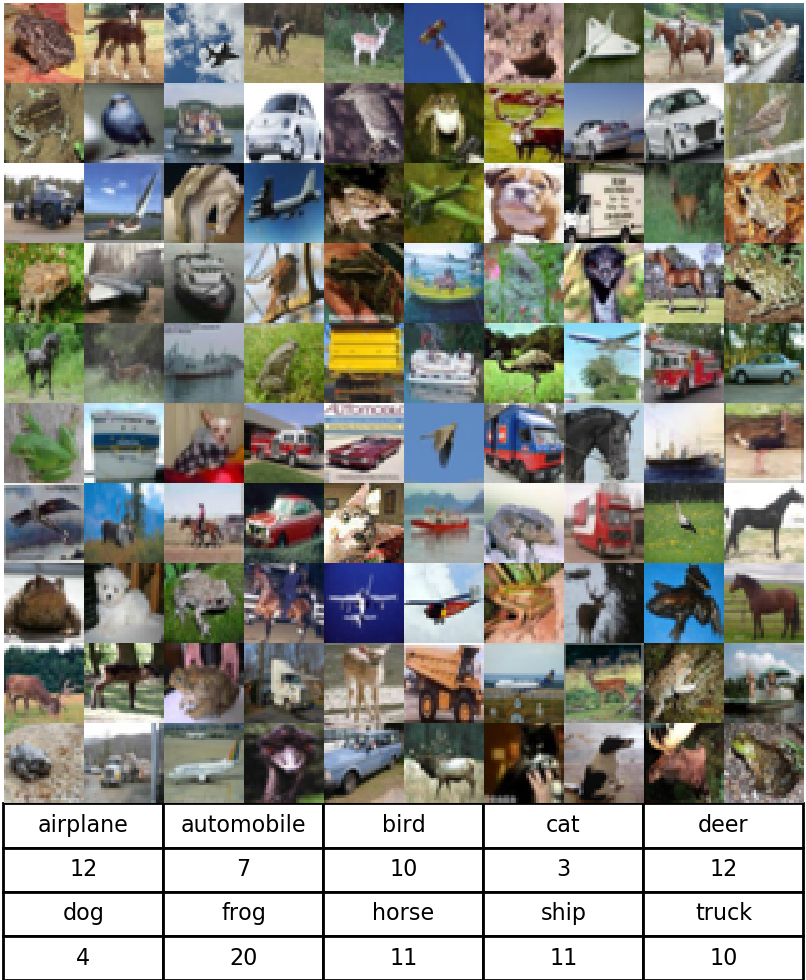}  
		\caption{VAAL - stage 2}
		\label{fig:VAA-3k}
	\end{subfigure}
	\begin{subfigure}{.33\textwidth}
		\centering
		\includegraphics[width=0.9\linewidth]{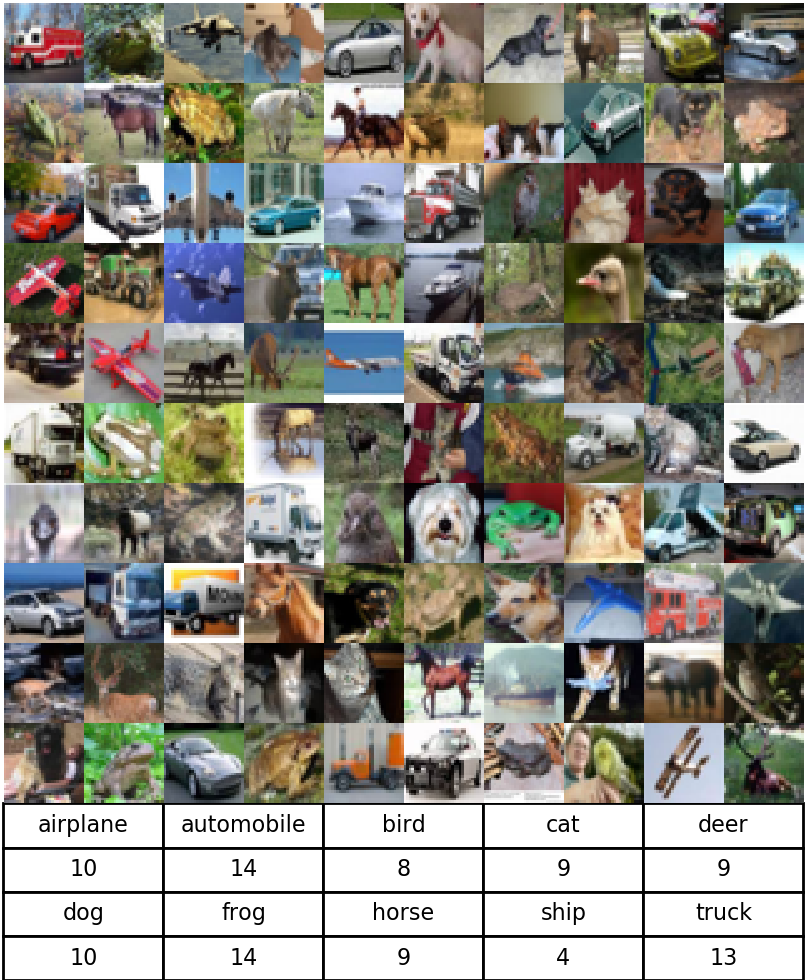}  
		\caption{VAAL - stage 3}
		\label{fig:VAA-4k}
	\end{subfigure}
	\begin{subfigure}{.33\textwidth}
		\centering
		\includegraphics[width=0.9\linewidth]{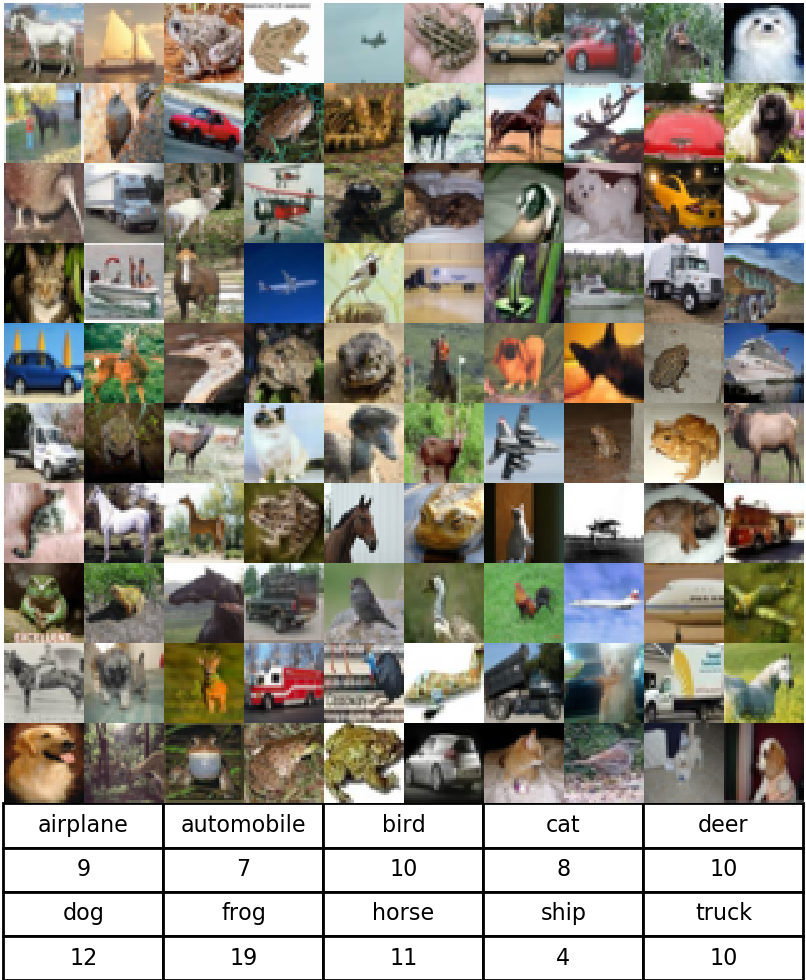}  
		\caption{TA-VAAL - stage 1}
		\label{fig:TA-VAAL-3K}
	\end{subfigure}
	\begin{subfigure}{.33\textwidth}
		\centering
		\includegraphics[width=0.9\linewidth]{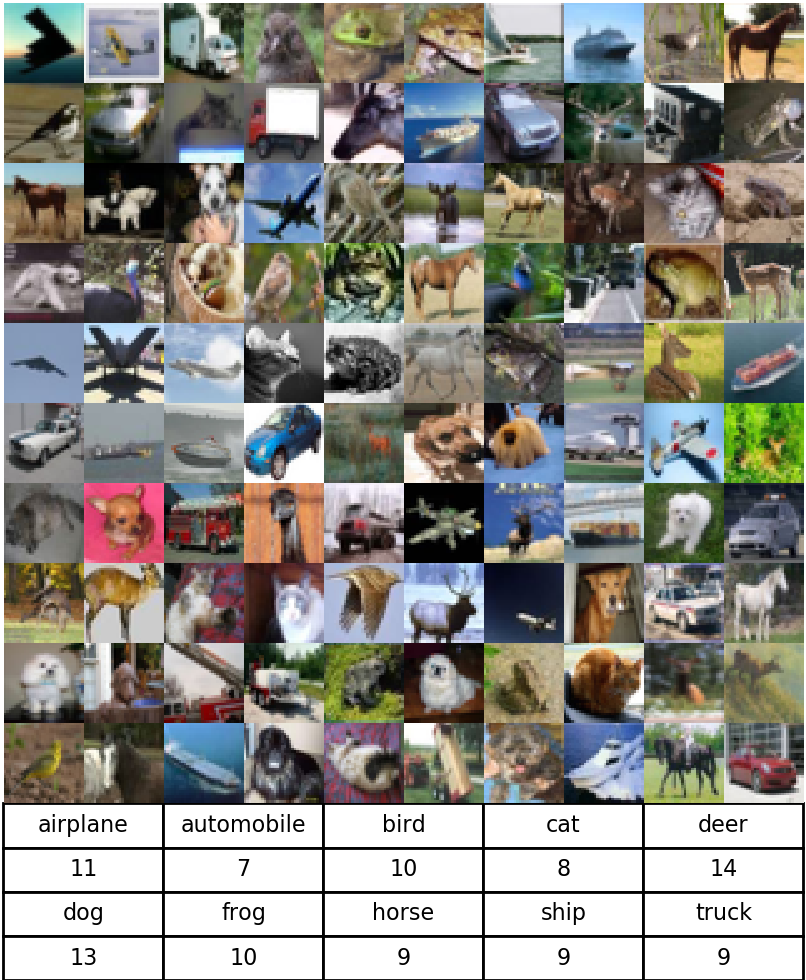}  
		\caption{TA-VAAL - stage 2}
		\label{fig:TA-VAAL-4K}
	\end{subfigure}
	\begin{subfigure}{.33\textwidth}
		\centering
		\includegraphics[width=0.9\linewidth]{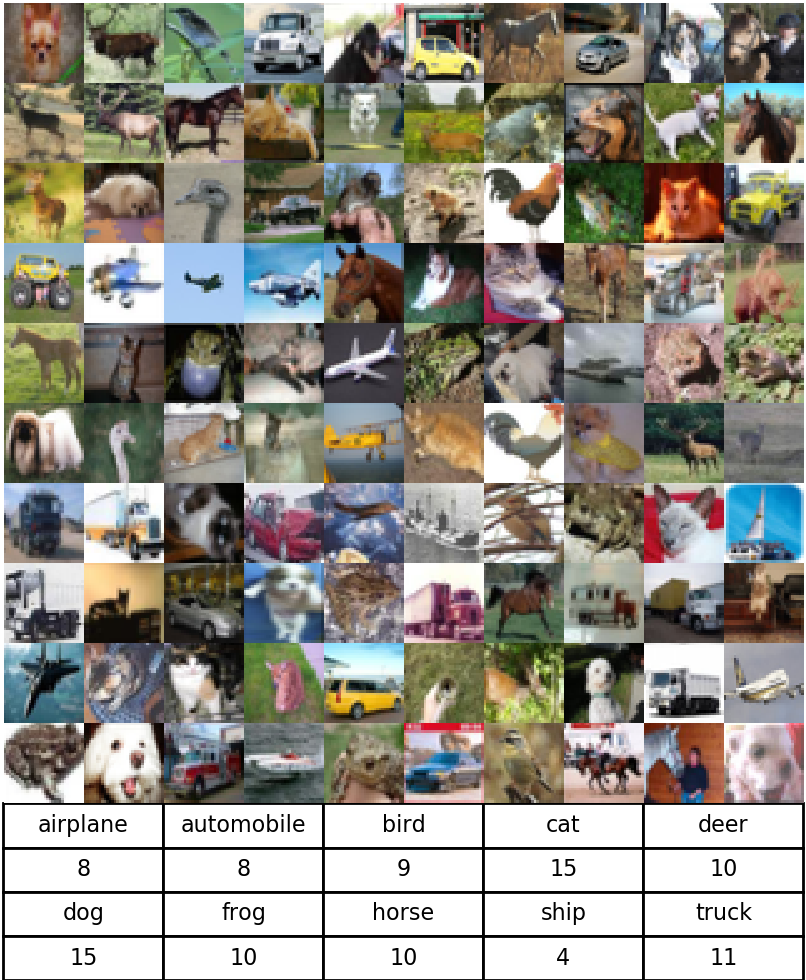}  
		\caption{TA-VAAL - stage 3}
		\label{fig:TA-VAAL-5K}
	\end{subfigure}
	\caption{Example images of active samples with other comparison methods on the modified CIFAR10 dataset with imbalance ratio $\times100$. Among budget 1k data points, top 100 data samples are displayed. The table represents the number of class counts.} 
	\label{fig:CIFAR10}
\end{figure*}

Figure~\ref{fig:CIFAR10} shows some selected images at each stage of different active learning methods, corresponding to the result of Figure 3(a) in the main paper.

	\section*{Acknowledgments}
	
	This work  was supported in part by a grant of the Korea Health Technology R\&D Project through the Korea Health
	Industry Development Institute (KHIDI), funded by the Ministry of Health \& Welfare, Republic of Korea (grant number : HI18C0316) and
	in part by the 2020 Research Fund(1.200033.01) of UNIST(Ulsan National Institute of Science and Technology).
		
{\small
\bibliographystyle{ieee_fullname}
\bibliography{reference}
}

\end{document}